\RequirePackage{fix-cm}
\documentclass{article}      
\usepackage{graphicx}
\usepackage[misc]{ifsym} 
\usepackage{mathptmx} 
\usepackage{latexsym}
\usepackage{rotating}

\usepackage[utf8]{inputenc}
\usepackage[T1]{fontenc}
\usepackage{hyperref}

\usepackage{natbib}

\usepackage{pgfplots}
\pgfplotsset{width=7cm,compat=1.8}
\usepackage{multirow}
\usepackage{xcolor}
\usepackage{leipzig}

\usepackage{rotating}

\usepackage{gb4e}

\noautomath

\makeatletter
\patchcmd{\@footnotetext}{\setcounter{fnx}{0}}{}{}{}
\apptocmd{\@footnotetext}{
    \@noftnotetrue
}{}{}
\@ifpackageloaded{bidi}{%
    \patchcmd{\@LTRfootnotetext}{\setcounter{fnx}{0}}{}{}{}
    \apptocmd{\@LTRfootnotetext}{
        \@noftnotetrue
    }{}{}
} 
\makeatother
\usepackage{multicol}
\usepackage{tikz-dependency}
\tikzset{/depgraph/.cd,/depgraph/.search also = {/tikz},
baseline=-0.6ex, inner sep=-0.1cm, edge horizontal padding=3pt, edge unit distance=1.8ex}

\usepackage{float}
\usepackage{array}

\usepackage{adjustbox}
\usepackage{listings}
\usepackage{authblk}

\definecolor{light}{rgb}{0.5, 0.5, 0.5}
\def\light#1{{\color{light}#1}}

\newcolumntype{L}[1]{>{\raggedright\let\newline\\\arraybackslash\hspace{0pt}}m{#1}}

\begin{document}
\title{Resources for Turkish Dependency Parsing: Introducing the BOUN Treebank and the BoAT Annotation Tool
}


\author[1]{Utku T\"{u}rk}
\affil[1]{Department of Linguistics, Bo\u{g}azi\c{c}i University}
\affil[2]{Department of Computer Engineering, Bogazici University}
\author[1]{Furkan Atmaca}
\author[2]{\c{S}aziye Bet\"{u}l \"{O}zate\c{s}}
\author[2]{G\"{o}zde Berk}
\author[1]{Seyyit Talha Bedir} 
\author[2]{Abdullatif K\"{o}ksal}
\author[1]{Balk{\i}z \"{O}zt\"{u}rk Ba\c{s}aran}
\author[2]{Tunga G\"{u}ng\"{o}r}
\author[2]{Arzucan \"{O}zg\"{u}r}



\maketitle

\begin{abstract}
In this paper, we introduce the  resources that we developed for Turkish dependency parsing, which include a novel manually annotated treebank (BOUN Treebank), along with  the guidelines we adopted, and a new annotation tool (BoAT). The manual annotation process we employed was shaped and implemented by a team of four linguists and five Natural Language Processing (NLP) specialists. Decisions regarding the annotation of the BOUN Treebank were made in line with the Universal Dependencies (UD) framework as well as our recent efforts for unifying the Turkish UD treebanks through manual re-annotation. To the best of our knowledge, BOUN Treebank is the largest Turkish treebank. It contains a total of 9,761 sentences from various topics including biographical texts, national newspapers, instructional texts, popular culture articles, and essays. In addition, we report the parsing results of a state-of-the-art dependency parser obtained over the BOUN Treebank as well as two other treebanks in Turkish. Our results demonstrate that the unification of the Turkish annotation scheme and the introduction of a more comprehensive treebank lead to improved performance with regards to dependency parsing. 
\end{abstract}




\section{Introduction}
\label{intro}

The field of Natural Language Processing (NLP) has seen an influx of various treebanks following the introduction of the treebanks in \citet{Marcus1993}, \citet{LeechGarside1991}, and \citet{Sampson1995}. These treebanks paved the way for today's ever-growing NLP framework, consisting of NLP applications, treebanks, and tools. Even though the value of a treebank cannot be judged solely by its number of sentences, previous research has shown that the size of a treebank may affect its utility in downstream NLP tasks \citep{foth2014size}. Among the many languages with a growing treebank inventory, Turkish has been one of the less fortunate languages. The latest version\footnote{UD version 2.7. Available at http://hdl.handle.net/11234/1-3424} of the Turkish IMST-UD Treebank is currently ranked as the 76\textsuperscript{th} treebank out of 183 treebanks in terms of the number of annotated sentences in the Universal Dependencies (UD) project \citep{nivre2016universal}. As of the UD version 2.7, the UD project includes 183 treebanks and the largest of them, the UD German-HDT Treebank, consists of 190,000 sentences \citep{borges-volker-etal-2019-hdt}. Turkish has posed an enormous challenge for NLP studies due to its complex network of inflectional and derivational morphology, as well as its highly flexible word order. One of the first attempts to create a structured treebank was initiated in the studies of \citet{Atalay2003} and \citet{Oflazer2003}. Following these studies, many more Turkish treebanking efforts were introduced \citep[][among others]{Megyesietal2010,Sulgeretal2013,sulubacak-etal-2016-universal}. However, most of these efforts contained a small volume of Turkish sentences, and some of them were re-introduced versions of already existing treebanks in a different annotation scheme. 

This paper aims to contribute to the limited NLP resources in Turkish by annotating a part of a brand new corpus that has not been approached with a syntactic perspective before, namely the Turkish National Corpus (henceforth TNC) \citep{Aksanetal2012}. TNC is an online corpus that contains 50 million words. The BOUN Treebank, which is introduced in this paper, includes 9,761 sentences extracted from five different text types in TNC, i.e. essays, broadsheet national newspapers, instructional texts, popular culture articles, and biographical texts. These sentences have not been introduced within a treebank previously.  
We manually annotated the syntactic dependency relations of the sentences following the up-to-date UD annotation scheme. 

Through a discussion of the annotation decisions made in the creation of the BOUN Treebank, we present our take on the annotation of Turkish data, including the challenges that the copular clitic, embedded constructions, compounds, and lexical cases pose. Turkish treebanking studies present an inconsistent picture in the annotation of such constructions, even though these linguistic phenomena are observed and studied extensively within Turkish linguistic studies.

In addition, we present a new annotation tool that integrates a tabular view, a hierarchical tree structure, and extensive morphological editing. We believe that other agglutinative languages that offer challenging morphological problems may benefit from this tool due to its ability to split and/or merge words and tokens in a sentence while rearranging the information regarding each word/token automatically, such as the word/token ID. This feature is crucial for the annotation process, since pre-processing of sentences may split the words and tokens erroneously.  

 Lastly, we report the results of an NLP task, namely dependency parsing, where we made parsing experiments on the newly introduced BOUN Treebank together with previous Turkish treebanks. 
 The results show that increasing the size of the training set has a positive  effect on the parsing success for Turkish. We observe that using the UD annotation scheme more faithfully and in a unified manner within Turkish UD treebanks offers an increase in the UAS (Unlabeled Attachment Score) F1 and LAS (Labeled Attachment Score) F1 scores. We also report individual parsing scores for different text types within our new treebank. 

This paper is organized as follows: In Section \ref{sec:turkish}, we briefly explain the morphological and syntactic properties of Turkish. In Section \ref{sec:litrev}, we present an extensive review of previous treebanking efforts in Turkish and locate them with regards to each other in terms of their use and their aim. In Section \ref{sec:treebank}, we report the details of the BOUN Treebank and our annotation process including the morphological and syntactic decisions. We lay out our tool BoAT in Section \ref{sec:tool}. In Section \ref{sec:exp}, we report our experiments and their results. In Section \ref{sec:conclusion}, we present our conclusions and discuss the implications of our work.

\section{Turkish} \label{sec:turkish}
Turkish is a Turkic language spoken mainly in Asia Minor and Thracia with approximately 75 million native speakers. As an agglutinative language, Turkish makes excessive use of morphological concatenation. According to \citet{wals-22}, a Turkish verb may have up to 8-9 inflectional categories per word, such as number, tense, or person marking. This number is about twice of the average of the maximum number of inflectional categories in the other 145 languages covered in  \citet{wals-22}. The number of morphological categories increases further when considering derivational processes. \citet{Kapan2019} states that Turkish words may host up to 6 different derivational affixes at the same time. The complexity of morphological analysis, however, is not limited to the sheer number of inflectional and derivational affixes. In addition to such affixes, allomorphies, vowel harmony processes, elisions, and insertions create an arduous task for researchers in Turkish NLP. \autoref{table:saketalAlin} lists the possible morphological analyses of the surface word \textit{al{\i}n}. The table shows that despite the shortness of the word, the morphological analysis is toilsome; and even such a short item may be parsed to have different possible roots.

\begin{sidewaystable}[]
\centering
\caption{Possible morphological analyses of the word \textit{al{\i}n} from \citet{sak:disamb:2008}. The symbol `\&' indicates derivational morphemes (originally `-', changed for clarity here), and `+' indicates inflectional morphemes. The strings between these symbols and the square-bracketed feature represent the phonology of the suffix. Upper case within a suffix means that the sound is phonologically conditioned. `H' stands for the archiphonemic high vowel. `N' stands for the allomorphy between the alveolar nasal and the lack of it. `Y' represents the allomorphy between the palatal glide and the lack of it.} 
\label{table:saketalAlin}     
\begin{tabular}{llllp{1.8cm}}
\hline\noalign{\smallskip}
Root & \begin{tabular}[c]{@{}l@{}}Category\\ of the root\end{tabular} & Features & Gloss & Translation \\
\noalign{\smallskip}\hline\noalign{\smallskip}
\textit{al{\i}n} & {[}Noun{]} & +{[}A3sg{]}+{[}Pnon{]}+{[}Nom{]}  & forehead & 'forehead' \\
\textit{al} & {[}Noun{]} & +{[}A3sg{]}+Hn{[}P2sg{]}+{[}Nom{]} & red-\textsc{poss} & 'your red color' \\
\textit{al} & {[}Adj{]} & \&{[}Noun{]}+{[}A3sg{]}+Hn{[}P2sg{]}+{[}Nom{]} & red-\textsc{poss} & 'your red color'\\
\textit{al} & {[}Noun{]} & +{[}A3sg{]}+{[}Pnon{]}+NHn{[}Gen{]} & red-\textsc{gen} & 'belonging to the color red'\\
\textit{al} & {[}Adj{]} & \&{[}Noun{]}+{[}A3sg{]}+{[}Pnon{]}+NHn{[}Gen{]} & red-\textsc{gen} & 'belonging to the color red'\\
\textit{al{\i}n} & {[}Verb{]} & +{[}Pos{]}+{[}Imp{]}+{[}A2sg{]} & offend-\textsc{2sg}.\textsc{imp} & 'Get offended!'\\
\textit{al} & {[}Verb{]} & +{[}Pos{]}+{[}Imp{]}+YHn{[}A2pl{]} & take-\textsc{2sg}.\textsc{hnr}-\textsc{imp} & '(Please) take it!'\\
\textit{al} & {[}Verb{]} & \&Hn{[}Verb+Pass{]}+{[}Pos{]}+{[}Imp{]}+{[}A2sg{]} & take-\textsc{pass}[\textsc{2sg}] & 'Get taken!'\\
\noalign{\smallskip}\hline
\end{tabular}
\end{sidewaystable}

With respect to syntactic properties, Turkish has a relatively free word order, which is constrained by discourse elements and information structure \citep{Taylan1986, Hoffman1995, kural1997postverbal, Issever2003, kornfilt2005asymmetries, Ozturk2008, ozturk2013postverbal,ozsoy2019word}. Even though SOV is the base word order, other permutations are highly utilized, as exemplified in \autoref{ordering}.\footnote{Conventions used in the paper are as follows: 
\textsc{1} = first person, \textsc{2} = second person, \textsc{3} = third person, \makebox{\textsc{abl} = ablative}, \textsc{acc} = accusative, \textsc{aor} = aorist, \textsc{caus} = causative, \textsc{cl} = classifier, \textsc{com} = comitative, \textsc{cond} = conditional, \textsc{cop} = copula, \textsc{cvb} = converb, \textsc{dat} = dative, \textsc{emph} = emphasis, \textsc{fut} = future, \textsc{gen} = genitive, \makebox{\textsc{hnr} = honorific}, \makebox{\textsc{imp} = imperative}, \textsc{loc} = locative, \textsc{neg} = negative, \textsc{nmlz} = nominalizer, \textsc{pass} = passive, \makebox{\textsc{pl} = plural}, \makebox{\textsc{poss} = possessive}, \textsc{prog} = progressive, \textsc{pst} = past, \textsc{q} = question particle, \textsc{sg} = singular. The dash symbol (\texttt{-}) in linguistics examples marks morpheme boundary, the equal sign (\texttt{=}) is used when the morpheme attached to a base is a clitic. The tilde \textit{$\sim$} is used to indicate partial replication. The asterisk \texttt{*} at the beginning of a sentence shows the sentence's ungrammaticality, and the percentage symbol (\texttt{\%}) shows the marginal acceptability of the sentence. Additionally, we presented the analytic words within a box when they are segmented for annotation.} The percentages were determined by \citet{SlobinBever1982} from 500 utterances of spontaneous speech. We also report word order percentages acquired from the BOUN Treebank in \autoref{table:wordorder} and \autoref{table:wordorder2} in Section \ref{appendix:wordorder}. These permutations are stemmed from processes including topicalization, focusing, and backgrounding. Contributing new or old information may also affect the place of a constituent, that is, new information may be placed closer to the verb and is always in pre-verbal position, whereas old information may surface both in pre-verbal and post-verbal positions. Another aspect that affects the word order is definiteness and specificity. Indefinite subjects and objects can typically surface in the immediately pre-verbal position.

\begin{exe}
\ex \label{ordering}
\begin{xlist}
\ex 
\gll Fatma Ahmet-i g{\"{o}}r-d{\"{u}}.  \textnormal{(SOV 48\%)}\\ 
Fatma Ahmet-\textsc{acc} see-\textsc{pst}\\
\glt `Fatma saw Ahmet.'
\ex Ahmet'i Fatma g{\"{o}}rd{\"{u}}. (OSV 8\%) 
\ex Fatma g{\"{o}}rd{\"{u}} Ahmet'i.  (SVO 25\%)
\ex Ahmet'i g{\"{o}}rd{\"{u}} Fatma. (OVS 13\%)
\ex G{\"{o}}rd{\"{u}} Fatma Ahmet'i.  (VSO 6\%)
\ex G{\"{o}}rd{\"{u}} Ahmet'i Fatma. (VOS <1\%) \hfill \citep[adapted from][]{Hoffman1995}
\end{xlist}
\end{exe}

As for the case system, every argument in a sentence needs to host a case according to its syntactic role, semantic contribution, or the lexical selection of the phrasal head \citep{taylan2015}. These groupings, however, are not clear cut and there is not always a one-to-one correspondence between cases and their roles. 

Moreover, Turkish is a pro-drop language in which the subject can be elided when it is retrievable from the given discourse \citep{Kornfilt1984,Ozsoy1988}. Overt subjects are used only to convey certain discourse and/or pragmatic effects, such as a change in context or focus. However, the subject is also retrievable from the agreement marker on the verb. In addition to these properties, Turkish is also a null object language, even though the language does not have an overt agreement marker available for this process \citep{Ozturk2006}. If the object of a sentence is retrievable from the given discourse, speakers may omit the object without any overt marking on the verb. The final issue with Turkish syntax lies in the fact that it frequently makes use of nominalization processes for embedded clauses \citep{goksel2005}. With certain nominalizer suffixes, the embedded sentences may function as an adverbial, an adjectival, or a nominal.

\section{Previous Turkish Treebank Initiatives}\label{sec:litrev}

The initial groundwork for Turkish treebanks was laid in \citet{Atalay2003} and \citet{Oflazer2003} following the studies on treebanks for languages such as English, German, Dutch, and many more \citep{LeechGarside1991,Marcus1993,Sampson1995,brants2002tiger,beek2002alpino}.  The first of its kind, the METU-Sabanc{\i} Treebank (MST) consists of 5,635 sentences, a subset of the METU corpus that reportedly includes 16 different text types such as newspaper articles and novels \citep{Say2002}. \citet{Oflazer2003} encoded both morphological complexities and syntactic relations. Due to the productive use of derivational suffixes, they explicitly spelled out every inflection and derivation within a word. As for the syntactic representation, \citet{Atalay2003} used a dependency grammar in order to bypass the problem of constituency in Turkish, which arises from the relatively free word order of the language.

Branching off the work of \citet{Atalay2003} and \citet{Oflazer2003}, a small treebank with the name of ITU Validation set for MST was introduced. It contains 300 sentences from 3 different genres. The treebank was introduced as a test set for MST in the CoNLL 2007 Shared Task \citep{eryigit2007itu}. The treebank was annotated by two annotators using a cross-checking process. Following this work, MST was re-annotated by \citet{sulubacak2016imst} from ground up with revisions made in syntactic relations and morphological parsing. The latest version was renamed as the ITU-METU-Sabanc{\i} Treebank (IMST). Due to certain limitations, \citet{sulubacak2016imst} employed only one linguist and several NLP specialists. The annotation process was arranged in such a way that there was no cross-checking between the works of the annotators. Moreover, inter-annotator agreement scores, details regarding the decision process among annotators, and the adjudication process have not been reported. Nevertheless, this re-annotation solved many issues regarding MST by proposing a new annotation scheme. Even though problems such as semantic incoherence in the usage of annotation tags and ambiguous annotation were resolved to a great extent, the non-communicative nature of the annotation process led to a handful of inconsistencies. 

The inconsistencies in IMST were also carried over to IMST-UD, which utilizes automatic conversions of the tags from IMST to the UD framework \citep{sulubacak-etal-2016-universal}. Mappings of syntactic and morphological representations were also included. Consequently,  IMST-UD was made more explanatory and clear thanks to the systematically added additional dependencies. While  IMST had 16 dependency relations, 47 morphological features, and 11 part of speech types, IMST-UD upped these numbers to 29, 66, and 14, respectively. Yet, the erroneous dependency tagging resulting from morpho-phonological syncretisms lingered long after the publication of the treebank. Moreover, no post-editing effort has been reported. There have been four updates since the first release of the IMST-UD treebank, but there are still mistakes that can be corrected through a post-editing process, such as the punctuation marks tagged as roots, reversed head-dependent relations, and typos in the names of syntactic relations.

Apart from the treebanks originating from MST, many other treebanks have emerged. Some of these treebanks can be grouped under the class of \textit{parallel treebanks}. The first of these parallel treebanks is the Swedish-Turkish Parallel Treebank (STPT). \citet{Megyesietal2008} published their parallel treebank containing 145,000 tokens in Turkish and 160,000 tokens in Swedish. Following this work, \citet{Megyesietal2010} published the Swedish-Turkish-English Parallel Treebank (STEPT). This treebank includes 300,000 tokens in Swe-dish, 160,000 tokens in Turkish, and 150,000 tokens in English. All the treebanks utilized the same morphological and syntactical parsing tools. For Swedish morphology, the Trigrams`n'Tags tagger \citep{Brants2000} trained on Swedish \citep{Megyesi2002} was used. On the other hand, Turkish data were first analyzed using the morphological parser 
in \citet{Oflazer1994}, and its accuracy was enhanced through the morphological disambiguator
proposed in \citet{YuretTure2006}. The Turkish and Swedish treebanks were annotated using the MaltParser \citep{nivre2007maltparser} that was trained with the Swedish treebank Talbanken05 \citep{Nivreetal2006Tal} and MST \citep{Oflazer2003}, respectively.

Another parallel treebank introduced for Turkish is the Turkish PUD Treebank, which adopts the UD framework. The Turkish PUD Treebank was published as part of a collaborative effort, the CoNLL 2017 Shared Task on Multilingual Parsing from Raw Text to Universal Dependencies \citep{zeman-EtAl:2017:K17-3}. Sentences for this collaborative treebank were drawn from newspapers and Wikipedia. The same 1,000 sentences were translated into more than 40 languages and manually annotated in line with the universal annotation guidelines of Google. After the annotation, the Turkish PUD Treebank was automatically converted to the UD style.

Moreover, there are three treebanks that consist of informal texts. One such treebank was introduced by \citet{Pamay2015} under the name of ITU Web Treebank (IWT). In IWT, non-canonical data were included such as the usage of punctuations in emoticons, abbreviated writing such as \textit{kib} that stands for \textit{kendine iyi bak} (take care of yourself), and non-standard writing conventions as in \textit{saol} instead of \textit{sa\u{g}ol} (thanks). Later on, the UD version of IWT was also introduced \citep{sulubacak2018implementing}. Another web treebank has recently been presented by \citet{Kayadelen2020}, which is larger than the previous Turkish treebanks in terms of word count, but still smaller than the BOUN Treebank that we introduce in this paper. \citet{Kayadelen2020} used a set of dependency labels similar to the UD framework. However, they diverge from the UD framework in certain issues such as postpositions, indirect objects, and oblique arguments. The Turkish-German Code-Switching Treebank \citep{CetinogluColtekin2016} is another treebank, in which they did not use formal texts. The Turkish-German Code-Switching Treebank consists of bilingual conversation transcriptions as well as their morphological and syntactic annotation. This treebank includes 48 unique conversations and 2,184 Turkish-German bilingual sentences that have been annotated with respect to the language in use.

There is also one grammar book-based treebank introduced in \citet{Coltekin2015}. The Grammar Book Treebank (GB) is the first UD attempt in Turkish treebanking. In this treebank, data were collected from a reference grammar book for Turkish written by \citet{goksel2005}. It includes 2,803 items that are either sentences or sentence fragments from the grammar book. It utilized TRMorph \citep{Coltekin2010} for morphological analyses and the proper morphological annotations were manually selected amongst the suggestions proposed by TRMorph. The sentences were manually annotated in the native UD-style.

In addition to these treebank initiatives, we recently started our unifying efforts in the syntactic annotation scheme in Turkish treebanking. We manually corrected the syntactic annotations in the Turkish PUD and IMST-UD treebanks \citep{turk-etal-2019-improving,turk-etal-2019-turkish}. In these works, we selected the treebanks that were not annotated natively in the UD style and unified the annotation scheme. This process improved the UAS score for the IMST-UD Treebank from 72.49 to 75.49 and caused only a 0.9 point decrease in the LAS score (from 66.43 to 65.53) in our experiments with the Standford's neural dependency parser \citep{dozat2017stanford}, despite the number of unique dependency tags increasing from 31 to 40 with the newly included dependency types \citep{turk-etal-2019-improving}. On the other hand, there was a decrease in the parsing accuracy for the re-annotated version of the PUD Treebank in terms of the attachment scores. While the parser achieved an UAS score of 79.52 and a LAS score of 73.81 on the previous version of the PUD Treebank, its attachment scores for the re-annotated version were 78.70 UAS and 70.01 LAS \citep{turk-etal-2019-turkish}. We want to note that, we used 5-fold cross validation for the evaluation of the PUD Treebank due to its small size. In each fold, the parser had only 600 sentences for training, and 200 sentences were used as the development set. The evaluation was done on the remaining 200 sentences. The small size of the PUD Treebank, which was originally used only for evaluation purposes (not for training) in the CoNLL 2017 Shared Task \citep{zeman-EtAl:2017:K17-3}, renders the results less reliable. Following these studies, with the annotation scheme we unified, we  manually annotated the BOUN Treebank, which we present in this paper. In \autoref{table:litrev}, we present basic statistics about the BOUN Treebank and compare it to the previous monolingual Turkish treebanks. If both UD and non-UD versions are available for a treebank, we only included the UD version in the table.\footnote{UD version number of these treebanks is 2.7. Turkish PUD version 2.7 is our re-annotated version.}

\begin{table}[hbt!]
\centering
\caption{Comparison of the BOUN Treebank to previous monolingual Turkish treebanks.} 
\label{table:litrev} 
\begin{tabular}{lrrrrrrr}
\hline\noalign{\smallskip}
 & IMST-UD & IWT-UD & GB    & PUD     & BOUN    \\
\noalign{\smallskip}\hline\noalign{\smallskip}
Num. of sentences     & 5,635    & 5,009   & 2,880  & 1,000  & 9,761  \\
Num. of tokens      & 56,396   & 44,463  & 16,803 & 16,536  & 121,214 \\
Avg. token count per sent. & 10.01 & 8.88 & 5.83 & 16.53  & 12.41\\
Avg. dep. arc length & 2.71 & 2.13 & 1.77 & 2.91  & 2.86 \\
Num. of unique POS tags      & 14       & 15      & 16     & 16       & 17      \\
Num. of unique features    & 66       & 54      & 79     & 59         & 56      \\
Num. of unique deps    & 32       & 28      & 41     & 40         & 41      \\
\noalign{\smallskip}\hline
\end{tabular}
\end{table}

\section[title]{The BOUN Treebank\footnotemark} \label{sec:treebank}

\footnotetext{Our treebank is available online at \url{https://github.com/UniversalDependencies/UD_Turkish-BOUN/}}

In this paper, we introduce a treebank that consists of 9,761 sentences which form a subset of the Turkish National Corpus (TNC) \citep{Aksanetal2012}.  TNC includes 50 million words from various text types, and encompasses sentences from a 20 year period between 1990 and 2009. The principles of the British National Corpus were followed in terms of the selection of the domains. \autoref{table:Aksan} shows the percentages of different domains and media used in TNC.

\begin{table}[hbt!]
\centering
\caption{Composition of the written component of TNC using words as the measurement unit, adapted from \citet{Aksanetal2012}.}
\label{table:Aksan}
\begin{tabular}{llll}
\hline\noalign{\smallskip}
Domain              & \%    & Medium                    & \%    \\
\noalign{\smallskip}\hline\noalign{\smallskip}
Imaginative         & 19    & Books                     & 58    \\
Social Science      & 16    & Periodicals               & 32    \\
Art                 & 7     & Miscellaneous published   & 5     \\
Commence/Finance    & 8     & Miscellaneous unpublished & 3     \\
Belief and Thought  & 4     & Written-to-be-spoken              & 2     \\
World Affairs       & 20    &                           &       \\
Applied Science     & 8     &                           &       \\
Nature Science      & 4     &                           &       \\
Leisure             & 14    &                           &       \\
\noalign{\smallskip}\hline
\end{tabular}
\end{table}

In our treebank, we included the following text types: essays, broadsheet national newspapers, instructional texts, popular culture articles, and biographical texts. Approximately 2,000 sentences were randomly selected from each of these registers. All of the selected sentences were written items and were not from the spoken medium. Our motivation for using these registers was to cover as many domains as possible using as few registers as possible, while not compromising variations in length, formality, and literary quality. TNC consists of 39 different registers, reported in \autoref{table:tncdetails2} in Section \ref{appendix:tncregisters}.\footnote{This table is retrieved from \url{https://www.tnc.org.tr/about-the-corpus/object/} on September 15, 2020.} The basic statistics for the BOUN Treebank and its different sections are provided in \autoref{table:tnc-stat}.

\begin{table}[hbt!]
\centering
\caption{Sentence and word statistics for the different sections of the BOUN Treebank. The difference between the number of tokens and words is due to multi-word expressions being represented with a single token, but with multiple words.}
\label{table:tnc-stat}
\begin{tabular}{cc}
\begin{tabular}{lrrr}
\hline

\bf Treebank & \bf N. of sent. & \bf N.of tokens & \bf N. of word forms \\
 \hline
Essays & 1,953 & 27,007 & 27,557 \\
Newspapers & 1,898  &  29,307 & 29,386\\
Instructional Texts & 1,976 & 20,442 & 20,625 \\
Popular Culture Articles & 1,962 & 21,067 & 21,263 \\
Biographical Texts & 1,972 & 23,391 & 23,553\\
\hline
\bf Total & \bf 9,761 & \bf 121,214 &\bf 122,384\\
\hline
\end{tabular}
\end{tabular}

\end{table}

Before the manual annotation of the BOUN Treebank, the sentences were first automatically annotated using an end-to-end parsing pipeline tool that parses raw texts to UD dependencies in CoNLL-U format with POS and morphological tagging information \citep{kanerva-EtAl:2018:K18-2}. The manual syntactic annotation of sentences were then performed on this automatically generated CoNLL-U versions of the corpus sentences.  
In the manual annotation process, we followed the UD syntactic relation tags. Before the annotation process started, we first reviewed the dependency relations in use within the UD framework. Upon reviewing the definitions, we created and annotated a list of unique sentences that we believe are representative of the UD dependency relations in Turkish. Later on, we compared our sentences for certain dependency relations with the examples from already existing Turkish UD treebanks. If our examples and the UD examples were not parallel, we first discussed whether or not our interpretation was correct. We then discussed whether or not there should be any inclusions to the UD guidelines. These discussion were also brought up within the UD community.

After settling on the definitions of the dependency relations, two Turkish native speaker linguists manually annotated the BOUN Treebank using our annotation tool that is presented in Section \ref{sec:tool}. Following the annotation process, two other linguists who did not participate in the manual annotation process cross-checked the syntactic annotations of the two linguists. When a problematic sentence or an inconsistency was encountered, discussions with regards to the sentence and related sentences were held among the team members. After a decision was made, the necessary changes were applied uniformly. 

In addition to the cross-checking process, we performed a partial double annotation in order to have a consistent annotation scheme before the annotation process of the BOUN Treebank started. For this purpose, the annotators performed an additional annotation task independently for the same set of 1,000 randomly selected sentences. The disagreements were discussed and resolved with the entire team of linguists and NLP specialists. The Cohen's Kappa measure of inter-annotator agreement for finding the correct dependency label of the relations is found to be 0.82. The unlabeled and labeled attachment scores between the annotations are 0.83 and 0.75, respectively.


\subsection{Levels of Annotation}

\subsubsection{Morphology}

Turkish makes use of affixation much more frequently than any other word-formation process. Even though it adds an immense complexity to its word level representation, patterns within the Turkish word-formation process allowed previous research to formulate morphological disambiguators that dissect word-level dependencies. One such work was introduced by \citet{sak2011resources}. Their morphological parser is able to run independently of any other external system and is capable of providing the correct morphological analysis with 98\% accuracy using contextual cues, such as the two previous tags. 

In the morphological annotation of the BOUN Treebank, we decided to use the morphological analyzer and disambiguator of \citet{sak2011resources}. For this purpose, the tokenized sentences were first given to the morphological parser. The output of the parser was converted to the corresponding UD features automatically. In rare cases where the morphological parser did not return a morphological analysis for a token, the morphological features column from the Turku pipeline \citep{kanerva-EtAl:2018:K18-2} for this token was used. The same operation was done for the lemmas of the tokens as well. 

Our preference for the morphological tagger of \citet{sak2011resources} instead of the morphological tagger of the Turku parsing pipeline \citep{kanerva-EtAl:2018:K18-2}, which we used for the automatic processing of the treebank in the first step, is due to their comparison in terms of the token-based accuracy,  and the feature-based recall,  precision, and f-measure metrics. After randomly selecting 50 words from every text type in the BOUN Treebank (a total of 250 unique tokens excluding punctuations for the five text types), we encoded the errors made by the morphological parsers. The results are shown in \autoref{table:SakvsUDpipe}. {\it Token Accuracy} column represents the token-based accuracy, namely the percentage of words for which correct morphological analyses are produced. {\it Recall} column represents the ratio of the number of correct morphological features to the number of morphological features in the gold standard. {\it Precision} column encodes the ratio of the number of correct morphological features to the total number of morphological features predicted by the morphological parser. The {\it F1-measure} column is the harmonic mean of precision and recall. Our scores align with the scores reported in the original study of \citet{sak2011resources}, even though their test set and our set here consist of different text types. While they only used newspaper corpora in the test set, we tested the parser using different text types including broadsheet national newspapers, essays, instructional texts, biographical texts, and popular culture articles. 

\begin{table}[hbt!]
\centering
\caption{ The performance of \citet{sak2011resources}'s and Turku pipeline's \citep{kanerva-EtAl:2018:K18-2} morphological taggers for BOUN Treebank.}
\label{table:SakvsUDpipe}

\begin{tabular}{lrrrr}
\hline\noalign{\smallskip}
Morphological Tagger & Token Accuracy & Recall & Precision & F1-measure \\
\noalign{\smallskip}\hline\noalign{\smallskip}
\citet{sak2011resources} & 0.91 & 0.94 & 0.95 & 0.94 \\
Turku pipeline & 0.82 & 0.89 & 0.83 & 0.86 \\
\noalign{\smallskip}\hline
\end{tabular}
\end{table}

The morphological parser of \citet{sak2011resources} does not provide morphological tags in UD format. So, we automatically converted its output to the UD format.
In  this process, we maximally used the morphological features from the UD framework. When there is no clear-cut mapping between the features that we acquired from the morphological parser of \citet{sak2011resources} and features proposed in the UD framework, we used the features previously suggested in the works of \citet{Coltekin2016},  \citet{Tyers2017}, and \citet{sulubacak2018implementing}. These features were already stated in the UD guidelines. \autoref{table:morphConversion} in Section \ref{appendix:conversion} shows the automatic conversion from the results of  \citet{sak2011resources}'s morphological disambiguator. As it is clear from the table, the depth of the morphological representation in \citet{sak2011resources} and that in the UD framework do not align perfectly, and there is no one-to-one mapping. For example, an output from \citet{sak2011resources} may include both \texttt{Narr} and \texttt{Past} features. In the automatic conversion, we would end up with \texttt{Tense=Past} twice and conflicting values for \texttt{Evident} feature. To resolve cases similar to these, we made use of simple rules that detect conflicting features due to our conversion and return appropriate features. Moreover, we used the morphological cues provided by the morphological parser to decide on the UPOS and lemma. All elements of our conversion and post-processing can be found on our Github page.\footnote{\url{https://github.com/boun-tabi/UD_docs}}

In our treebank, in addition to the words, we encoded the lexical and grammatical properties of the words as sets of features and values for these features. We also encoded the lemma of every word separately, following the UD framework. \autoref{table:conllu-file} shows an example sentence encoded with the CoNLL-U format. 

\begin{table}[hbt!]
\centering
\caption{ An example sentence from our treebank encoded in CoNLL-U format.}
\scriptsize
\label{table:conllu-file}

\tabcolsep=0.049cm
\begin{tabular}{lllllL{4cm}llll}
\hline\noalign{\smallskip}
\multicolumn{10}{l}{\# sent\_id = ins\_167}                                                                                                                    \\
\multicolumn{10}{l}{\# text = S\"{o}z\"{u} uzat{\i}p seni merakta b{\i}rakt{\i}m galiba.}                                                                                       \\
\multicolumn{10}{l}{\# trans = Probably, I beat around the bush and kept you in suspense.}                                                                                       \\
\it \light{ID} &\it  \light{FORM} &\it \light{LEMMA} &\it \light{UPOS} &\it \light{XPOS} &\it \light{FEATS} &\it \light{HEAD} &\it \light{DEPREL} &\it \light{DEPS} &\it \light{MISC} \\
1 & S\"{o}z\"{u}     & s\"{o}z    & NOUN  & Noun   & Case=Acc|Number=Sing|Person=3                                       & 2 & obj    & \_ & \_                            \\
2 & uzat{\i}p   & uza    & VERB  & Verb   & Polarity=Pos|VerbForm=Conv| Voice=Cau                                              & 5 & advcl  & \_ & \_                            \\
3 & seni     & sen    & PRON  & Pers   & Case=Acc|Number=Sing|Person=2                                       & 5 & obj    & \_ & \_                            \\
4 & merakta  & merak  & NOUN  & Noun   & Case=Loc|Number=Sing|Person=3                                       & 5 & obl    & \_ & \_                            \\
5 & b{\i}rakt{\i}m & b{\i}rak  & VERB  & Verb   & Aspect=Perf|Evident=Fh|Number=Sing| Person=1|Polarity=Pos|VerbForm=Fin| Tense=Past & 0 & root   & \_ & \_                            \\
6 & galiba   & galiba & ADV   & Adverb & \_                                                                  & 5 & advmod & \_ & SpaceAfter=No                 \\
7 & .        & .      & PUNCT & Punc   & \_                                                                  & 5 & punct  & \_ & SpacesAfter=\textbackslash{}n \\
\noalign{\smallskip}\hline
\end{tabular}
\end{table}

\subsubsection{Syntax}

In the BOUN Treebank, we decided to represent the relations amongst the parts of the sentences within a dependency framework. This decision has two main reasons. The main and the historical reason is the fact that the growth of Turkish treebanks has been mainly within the frameworks where the syntactic relations have been represented with dependencies \citep{Oflazer1994,Cetinoglu2009LFG}. The other reason is the fact that Turkish allows for phrases to be scrambled to pre-subject, post-verbal, and any clause-internal positions with specific constraints, which makes building constituency grammars quite difficult \citep{Taylan1984, Kural1992, Aygen2003, Issever2007}. With these in mind, we wanted to stick with the conventional dependency framework and use the recently rising UD framework.\footnote{For more information on the UD framework, see \url{https://universaldependencies.org/u/dep/index.html}. For our annotation guidelines, please see \url{https://github.com/boun-tabi/UD_docs/blob/main/_tr/dep/Turkish_deprel_guidelines.pdf}.} One of the main advantages of the UD framework is that it creates directly comparable sets of treebanks with regards to their syntactic representation due to its very nature.

By following the UD framework, we implicitly encode two different syntactic information for each dependent: the category of the dependent and the function of this dependent with regards to its syntactic head. This is due to the grouping of the dependency relations introduced by the UD framework. The selection of the syntactic dependency relation for each dependent is mainly based on the functional category of the dependent in relation to the head and the structural category of the head. In terms of the functional category of the dependent, the UD framework differentiates the core arguments of clauses, non-core arguments of clauses, and dependents of nominal heads. As for the category of the dependent, the UD framework makes use of a taxonomy that distinguishes between function words, modifier words, nominals, and clausal elements. In addition to this classification, there are some other groupings which may be listed as: coordination, multiword expressions, loose syntactic relation, sentential, and extra-sentential.\footnote{For the complete table of syntactic relations, please check \url{https://universaldependencies.org/u/dep/index.html}} \autoref{table:deprel} shows the dependency relations that we  employed in the BOUN treebank with their counts and percentages.

\begin{table}[hbt!]
\centering
\caption{The dependeny relation set of the BOUN Treebank.}
\label{table:deprel}
\begin{tabular}{lrrllrr}
\hline\noalign{\smallskip}
Relation Type  & Count & Percentage &  & Relation Type & Count & Percentage \\
\noalign{\smallskip}\hline\noalign{\smallskip}
\texttt{acl}            &   3,494  &2.85\% &  & \texttt{det}           &  4,938 &4.03\% \\
\texttt{advcl}          &   2,595  &2.12\% &  & \texttt{discourse}     &  381   &0.31\% \\
\texttt{advcl:cond}     &   269    &0.22\% &  & \texttt{dislocated}    &  28    &0.02\% \\
\texttt{advmod}         &   5,278  &4.31\% &  & \texttt{fixed}         &  12    &0.01\% \\
\texttt{advmod:emph}    &   1,724  &1.41\% &  & \texttt{flat}          &  2,039 &1.67\% \\
\texttt{amod}           &   7,869  &6.43\% &  & \texttt{goeswith}      &  4     &0.002\% \\
\texttt{appos}          &   506    &0.41\% &  & \texttt{iobj}          &  164   &0.13\% \\
\texttt{aux}            &   39     &0.03\% &  & \texttt{list}          &  40    &0.03\% \\
\texttt{aux:q}          &   269    &0.22\% &  & \texttt{mark}          &  117   &0.10\% \\
\texttt{case}           &   3,290  &2.69\%  &  & \texttt{nmod}          &  1,371 &1.12\% \\
\texttt{cc}             &   2,800  &2.29\% &  & \texttt{nmod:poss}     &  10,393&8.49\% \\
\texttt{cc:preconj}     &   134    &0.11\% &  & \texttt{nsubj}         &  8,499 &6.94\% \\
\texttt{ccomp}          &   1,512  &1.24\% &  & \texttt{nummod}        &  1,568 &1.28\% \\
\texttt{clf}            &   122    &0.1\%  &  & \texttt{obj}           &  7,381 &6.03\% \\
\texttt{compound}       &   2,381  &1.95\% &  & \texttt{obl}           &  12,015&9.82\% \\
\texttt{compound:lvc}   &   1,218  &1.0\%  &  & \texttt{orphan}        &  84    &0.07\% \\
\texttt{compound:redup} &   457    &0.37\% &  & \texttt{parataxis}     &  209   &0.17\% \\
\texttt{conj}           &   7,250  &5.92\% &  & \texttt{punct}         &  20,116&16.44\% \\
\texttt{cop}            &   1,289  &1.05\% &  & \texttt{root}          &  9,761 &7.97\% \\
\texttt{csubj}          &   546    &0.45\% &  & \texttt{vocative}      &  88    &0.07\% \\
\texttt{dep}            &   9      &0.01\% &  & \texttt{xcomp}         &  125   &0.01\% \\           
\noalign{\smallskip}\hline
\end{tabular}
\end{table}

Every dependency forms a relation between two segments within the sentence, building up to a non-binary and hierarchical representation of the sentence. In this way, nodes can have more than two children nodes and every node is accessible from the root node. This representation is exemplified in \autoref{ex:tncExampleSentence} using the sentence in \autoref{table:conllu-file}.

\begin{exe}
\ex \label{ex:tncExampleSentence}
\scalebox{.9}{\begin{dependency}
\begin{deptext}
S\"{o}z-\"{u} \& uza-t-{\i}p \& sen-i \& merak-ta \& b{\i}rak-t{\i}-m \& galiba \& . \\ word-\textsc{acc}  \& strech-\textsc{caus}-\textsc{nmlz} \& you-\textsc{acc}  \& curiosity-\textsc{loc}  \& leave-\textsc{pst}-\textsc{1sg}  \& probably  \& .\\
\end{deptext}
\deproot{4}{\textsc{root}}
\depedge{4}{6}{\textsc{advmod}}
\depedge{4}{7}{\textsc{punct}}
\depedge{4}{5}{\textsc{compound:lvc}}
\depedge{4}{3}{\textsc{obj}}
\depedge{4}{2}{\textsc{advcl}}
\depedge{2}{1}{\textsc{obj}}
\end{dependency}}
    \glt `Probably, I beat around the bush and kept you in suspense.'

\end{exe}

\subsection{Different Conventions Adopted in the Annotation Process}

In the annotation process of the BOUN Treebank, we stayed faithful to the UD main tag set and the previous conventions of Turkish annotation schemes for the most part. However, there were some instances where we diverged from these conventions or made the linguistic reasoning behind them more explicit. In this section, we provide the justifications of our linguistic decisions for 
these instances. Our decisions are in the same spirit of unifying the annotation scheme within Turkish UD treebanks, which was done in our previous works \citep{turk-etal-2019-improving,turk-etal-2019-turkish}. Our main concern is to reflect linguistic adequacy in the BOUN Treebank following the Manning's Law \citep{Nivre2017}. During all this work, we paid great attention to follow the previous discussion within the UD framework, such as the discussion on the copular clitic and the objecthood-case marking relation. In the following sections, we will first touch upon the issues where we believe the previous conventions in Turkish UD treebanking were erroneous according to UD. These issues include the annotation of the embedded sentences, the treatment of copular verb, the analysis of compounds, and the annotation of classifiers. Next, we will discuss the issue of objecthood and the case marking relation in Turkish, where we adopt a simpler analysis that has been used in other dependency grammars instead of the recently discussed UD alternatives.

\subsubsection{Annotation of Embedded Clauses} 

The first issue where we diverged from the previous annotation conventions is the annotation of embedded clauses. In the previous treebanks, the annotation of embedded clauses did not reflect the inner hierarchy that a clause by definition possesses. This is mostly due to the morphological aspect of the most common embedding strategy in Turkish: nominalization. Due to nominalization, embedded clauses in Turkish can be regarded as nominals since they behave exactly like nominals: They can be marked with an accusative case, can be substituted with any other nominal, and can carry genitive-possessive cases as person marking as shown in \autoref{ex:embeddedLing}. The embedded clause in the given sentence is shown with square brackets. The whole square bracket can be replaced with a simple noun, like \textit{otob{\"{u}}s} (bus), or a complex noun phrase like \textit{senin otob{\"{u}}s{\"{u}}n} (your bus) as in \autoref{ex:embeddedling2}.

\begin{exe}
    \ex \label{ex:embeddedLing}
    \gll \textnormal{[} Sen-in otob{\"{u}}s-{\"{u}} s{\"{u}}r-d{\"{u}}{\u{g}}-{\"{u}}n \textnormal{]}-{\"{u}} g{\"{o}}r-d{\"{u}}-m.\\
    [ you-\textsc{gen} bus-\textsc{acc} drive-\textsc{nmlz}-\textsc{poss} ]-\textsc{acc} see-\textsc{pst}-\textsc{1sg} \\
    \glt `I saw that you drove the bus.'
    \ex \label{ex:embeddedling2}
    \gll Sen-in otob{\"{u}}s-{\"{u}}n-{\"{u}} g{\"{o}}r-d{\"{u}}-m.\\
    you-\textsc{gen} bus-\textsc{poss}-\textsc{acc}  see-\textsc{pst}-\textsc{1sg} \\
    \glt `I saw your bus.'
\end{exe}

Due to these surface level morphological and syntactic similarities, previous Turkish treebanks in the UD framework, with the exception of the Grammar Book  Treebank \citep{Coltekin2015}, used dependency relation \texttt{obj} instead of \texttt{ccomp}, \texttt{nsubj} instead of \texttt{csubj}, \texttt{amod} instead of \texttt{acl}, and \texttt{advmod} instead of \texttt{advcl} to mark the relation of the embedded clause with the matrix verb. In our annotation process, we emphasized the clausal nature of these embedded sentences and their syntactic derivation by focusing on their internal structure reflecting the existence of a temporal domain in the embedded clause. For instance, \autoref{ex:embeddedLing} would be unsensical if we had the time adverb \textit{tomorrow}
 within the embedded clause. This ungramaticality is due to the tense information introduced by the nominalizer `-d{\"{u}}{\u{g}}' in the example sentence. If there were an adverb like {\it tomorrow} in an embedded clause marked with `-d{\"{u}}{\u{g}}', the previous annotation scheme would not be able to detect the ungrammaticality. However, our annotation scheme is able to detect this ungrammaticality.

The same argumentation applies to converbs, as well. Converbs are verbal elements of a non-finite adverbial clause \citep{goksel2005}. They may act as adverbial adjuncts or as discourse connectives. In the previous annotation processes of Turkish, they were annotated as \texttt{nmod}. The reason behind this annotation is again the fact that they behave like nominals; they may be marked with inflectional and derivational suffixes that normally nouns bear. Considering their clausal properties, such as their temporal domain, their ability to host a subject, an object, and a tense/aspect/modality information, we annotated them as \texttt{advcl} as 
in \autoref{ex:nmodtoadvcl}.\footnote{Throughout the paper, changes in the annotation convention introduced by us are shown with bold arcs, whereas the dashed arcs suggest previous annotations. The solid arcs represent unaltered dependencies. Every annotated tree that contains a bold arc in this paper is taken from previous Turkish Treebanks, that is either the IMST-UD Treebank or the Turkish PUD Treebank.}

\begin{exe}
\ex
        \scalebox{0.95}{\begin{dependency}
            \begin{deptext}[column sep=.3cm]
            Bira-lar-{\i} \& devir-dik-\c{c}e \& merak-{\i}m \& az-d{\i} \& . \\
            beer-\textsc{pl}-\textsc{acc} \& topple-\textsc{nmlz}-\textsc{cvb} \& curiosity-\textsc{1sg.poss} \& get.wild-\textsc{pstw} \& . \\
        \end{deptext}
            \deproot{4}{\textsc{root}}
            \depedge{2}{1}{\textsc{obj}}
            \depedge{4}{5}{\textsc{punct}}
            \depedge[edge style={ultra thick}]{4}{2}{\textsc{\textbf{advcl}}}
            \depedge[edge unit distance=3ex,edge style={densely dotted}]{4}{2}{\textsc{nmod}}
            \depedge{4}{3}{\textsc{nsubj}}
        \end{dependency}}
    \glt ``As I finish my beers, my curiosity peaked."
    \label{ex:nmodtoadvcl}
\end{exe}

In addition to the annotation of the whole embedded clause, dependents within the embedded clause were erroneously annotated in the previous Turkish annotation schemes. For example, an oblique of an embedded verb used to be attached to the root since the embedded verb is seen as a nominal, and not as a verb as in \autoref{ex:embedded_bad_1}.

       \begin{exe} 
            \ex \label{ex:embedded_bad_1} \scalebox{0.95}{
                    \begin{dependency}
                        \begin{deptext}[column sep=0.1cm]
T{\"{u}}nel-e \& gir-me-den \& {\"{o}}nce \& ge{\c{c}}-ti{\u{g}}-im \& \ldots \& bam\textasciitilde ba{\c{s}}ka \& .\\
tunnel-\textsc{dat} \& enter-\textsc{neg}-\textsc{nmlz} \& before \& pass-\textsc{nmlz}-\textsc{1sg} \& \ldots \& \textsc{emph}\textasciitilde different\& .\\
                    \end{deptext}
                    \deproot{6}{\textsc{root}}
                    \depedge[edge unit distance=0.6ex, edge below,edge style={densely dotted}]{6}{1}{\textsc{nmod}}
                    \depedge[edge style={ultra thick}]{2}{1}{\textsc{obl}}
                    \depedge{4}{2}{\textsc{advcl}}
                    \depedge{2}{3}{\textsc{case}}
                    \depedge{5}{4}{\textsc{acl}}
                    \depedge{6}{7}{\textsc{punct}}
                    \depedge{6}{5}{\ldots}
                    \end{dependency}
                    }
            \glt `The scenery that I passed before I entered the tunnel was completely different from here.'
        \end{exe}

Likewise, the genitive subjects of embedded clauses were wrongly marked as a possessive nominal modifier, whereas they are one of the obligatory elements of the embedded structures. This wrong annotation in the previous treebanks is due to the fact that Turkish makes use of genitive-possessive structure for marking the agreement in an embedded clause as in \autoref{ex:embedded_bad_2} \citep{goksel2005}. Despite the morphology, the word \emph{senin} here serves as the subject. \autoref{ex:nsubjproof} shows the causativized version of the embedded verb in \autoref{ex:embedded_bad_2}. When we causativize the subject of an intransitive verb, we expect the subject to be marked with an accusative case and act as a direct object. As seen in \autoref{ex:embedded_bad_2} and \ref{ex:nsubjproof}, the word \textit{sen} reflects the morphological reflex stemming from a syntactic voice change. Thus, it cannot be a modifier and it has to be an argument.
        \begin{exe}
                \ex \label{ex:embedded_bad_2}
        \begin{dependency}
            \begin{deptext}[column sep=.1cm]
           Sen-in \& de \& gel-me-n-i \& iste-r-di-m \& . \\
           you-\textsc{gen} \& too \& come-\textsc{nmlz}-\textsc{poss}-\textsc{acc} \& want-\textsc{aor}-\textsc{pst}-\textsc{1sg}\&. \\
        \end{deptext}
            \deproot{4}{\textsc{root}}
            \depedge{4}{3}{\textsc{ccomp}}
            \depedge{4}{5}{\textsc{punct}}
            \depedge[edge below, label style={below}]{1}{2}{\textsc{advmod:emph}}
            \depedge[edge unit distance=1ex, edge style={ultra thick}]{3}{1}{\textsc{\textbf{nsubj}}}
            \depedge[edge unit distance=2ex,edge style={densely dotted}]{3}{1}{\textsc{nmod:poss}}
        \end{dependency}
    \glt `I would have wanted you to come, as well.'
    
    \ex \label{ex:nsubjproof}
    \gll O-nun sen-i de getir-me-si-ni iste-r-di-m.\\
    he/she-\textsc{gen} you-\textsc{acc} too come.\textsc{caus}-\textsc{nmlz}-\textsc{poss}-\textsc{acc} want-\textsc{aor}-\textsc{pst}-\textsc{1sg}.\\
    \glt `I would have wanted him/her to bring you, as well.'
\end{exe} 

Due to the reasons explained above, in the annotation of embedded clauses we used the dependency relations that emphasize the clausal nature of the nominalized verbs, i.e., \texttt{csubj}, \texttt{ccomp}, \texttt{advcl}, instead of the dependency relations that emphasize the final product of the local derivations, i.e., \texttt{nsubj}, \texttt{obj}, \texttt{advmod}, respectively.

\subsubsection{Copular Clitic}

One inconsistent issue within the Turkish treebanks was the annotation of the copular clitics. Copular clitics attached to the verbal bases and nominal bases were treated differently although they are essentially the same as we will show below. While the copular clitics on verbal bases were not segmented, the copular clitics on nominal bases were segmented in previous Turkish treebanks. In this section, we will provide our analysis where we segment all copular clitics regardless of their bases. 

The Turkish copular clitic is the grammaticalized version of the verb "be" which can be indicated as \emph{i-}. This clitic \emph{i-} has three allomorphs in Turkish: (i) analytic \emph{i-}, (ii) suffixal \emph{-y}, and (iii) zero-marked (\O). The allomorphy of the analytic form is idiosyncratic, meaning the analytic copula form can be used in place the suffixal copula forms most of the time. The analytic form can surface if suffixes  \emph{-di} (\textsc{pst}), \emph{-se} (\textsc{cond}), and \emph{-ken} (\textsc{when} or \textsc{while}) come atop a verb that already hosts a TAM (Tense/Aspect/Modality) marker. The analytic form can also surface in nominal sentences that are marked for tense other than the aorist (\textit{-Ar/Ir}). However, the analytic form cannot surface with the suffix \textit{-mI{\c{s}}} (\textsc{prf}), except for its use with the aorist as in \textit{yapar imi\c{s}}, meaning {\it he or she used to do}. \autoref{ex:VerbalAnalyBe} and \autoref{ex:VerbalAnalyBe2} illustrate some examples of the analytic form.

\begin{exe}
\ex \begin{xlist}
\begin{multicols}{2}
            \ex \label{ex:VerbalAnalyBe} analytic \textsc{cop} \emph{i-} \\ \scalebox{1}{
                    \hspace{-2em}\begin{dependency}
                        \begin{deptext}
                        Okul-a \& var-acak \& i-di-m \& . \\
                        school-\textsc{dat} \& reach-\textsc{fut} \& be-\textsc{pst}-\textsc{1sg} \& .\\
                    \end{deptext}
                    \deproot{2}{\textsc{root}}
                    \depedge{2}{1}{\textsc{obl}}
                    \depedge{2}{3}{\textsc{cop}}
                    \depedge{2}{4}{\textsc{punct}}
                    \end{dependency}
                    }
            \glt \hspace{-2em}`I was going to arrive to school.'
\columnbreak            
            \ex \label{ex:VerbalAnalyBe2} analytic \textsc{cop} \emph{i-} \\ \scalebox{1}{
                    \begin{dependency}
                        \begin{deptext}
                        Okul-a \& var-acak \& i-ken \& . \\
                        school-\textsc{dat} \& reach-\textsc{fut} \& be-\textsc{when} \& . \\
                    \end{deptext}
                    \deproot{2}{\ldots}
                    \depedge{2}{1}{\textsc{obl}}
                    \depedge{2}{3}{\textsc{cop}}
                    \depedge{2}{4}{\textsc{punct}}
                    \end{dependency}
                    }
            \glt `I was about arrive to school. \ldots'
\end{multicols}            
\end{xlist}            
\end{exe}

When both the base and the copular verb surface as a single syntactic word indicated with a box in the following examples, either \textit{-y} (\autoref{ex:TurkishBeizero}, \ref{ex:TurkishBey2}) or {\O} (\autoref{ex:TurkishBeizeroy}, \ref{ex:TurkishBey3}) is used. The selection between the {\O} and \textit{-y} is governed by the phonological characteristics of the previous sound; if the previous segment is a consonant {\O} is used, otherwise \textit{-y} is used.

\begin{exe}
\ex%
\begin{xlist}
\begin{multicols}{2}            
             \ex \label{ex:TurkishBeizero} zero-marked \textsc{cop} (\O)\\\scalebox{0.85}{
                    \begin{dependency}
                        \begin{deptext}
                        Okul-a \& gel-ecek \& =ti-m \& . \\
                        school-\textsc{dat} \& come-\textsc{fut} \& =\textsc{pst}-\textsc{1sg} \&  . \\
                    \end{deptext}
                    \deproot{2}{\textsc{root}}
                    \depedge{2}{1}{\textsc{obl}}
                    \depedge{2}{3}{\textsc{cop}}
                    \depedge{2}{4}{\textsc{punct}}
                    \wordgroup{1}{2}{3}{verb}
                    \end{dependency}
                    }
            \glt `I was going to come to school.'
\columnbreak
            \ex \label{ex:TurkishBey2} zero-marked \textsc{cop} (\O)\\\scalebox{0.85}{
                    \begin{dependency}
                        \begin{deptext}
                        Okul-da \& {\"{o}}{\u{g}}retmen \& =di-m \& . \\
                        school-\textsc{loc} \& teacher \& =\textsc{pst}-\textsc{1sg} \& . \\
                    \end{deptext}
                    \deproot{2}{\textsc{root}}
                    \depedge{2}{1}{\textsc{obl}}
                    \depedge{2}{3}{\textsc{cop}}
                    \depedge{2}{4}{\textsc{punct}}
                    \wordgroup{1}{2}{3}{verb}
                    \end{dependency}
                    }
            \glt `I was a teacher in the school.'
\end{multicols}
\end{xlist} 

\ex \begin{xlist}
\begin{multicols}{2}
            \ex \label{ex:TurkishBeizeroy} suffixal \textsc{cop} \textit{-y}\\\scalebox{0.85}{
                    \begin{dependency}
                        \begin{deptext}
                        Okul-a \& gel-se \& =y-di-m \& . \\
                        school-\textsc{dat} \& come-\textsc{cond} \& =\textsc{cop}-\textsc{pst}-\textsc{1sg}\& .\\
                    \end{deptext}
                    \deproot{2}{\textsc{root}}
                    \depedge{2}{1}{\textsc{obl}}
                    \depedge{2}{3}{\textsc{cop}}
                    \depedge{2}{4}{\textsc{punct}}
                    \wordgroup{1}{2}{3}{verb}
                    \end{dependency}
                    }
            \glt `If I went to school.'
\columnbreak        
            \ex \label{ex:TurkishBey3} suffixal \textsc{cop} \textit{-y}\\\scalebox{0.85}{
                    \begin{dependency}
                        \begin{deptext}
                        Okul-da \& {\"{o}}{\u{g}}renci \& =y-se-m \& . \\
                        school-\textsc{loc} \& student \& =\textsc{cop}-\textsc{cond}-\textsc{1sg} \& .\\
                    \end{deptext}
                    \deproot{2}{\textsc{root}}
                    \depedge{2}{1}{\textsc{obl}}
                    \depedge{2}{3}{\textsc{cop}}
                    \depedge{2}{4}{\textsc{punct}}
                    \wordgroup{1}{2}{3}{verb}
                    \end{dependency}
                    }
            \glt `If I was a student in the school...'
\end{multicols}
\end{xlist}

\end{exe}

What is important for us is that the contribution of these copular clitics is the same for both nominal and verbal bases. In both cases, these copular clitics host the TAM information that cannot be carried by the base \citep{Goksel2001}. The TAM information itself also does not change according to the category of the stem. 

Additionally, the stress patterns of the clitics that attach to nominal and verbal bases are identical. Most of the verbs and common nouns are stressed in the final syllable. When they are marked with a copular clitic, instead of the final syllable which is the copular clitic, the preceding syllable is stressed \citep{Goksel2001}. This property as well applies regardless of the base the clitic attaches to.

In addition to these characteristics, the copular clitic also has a clitic-like behaviour when it co-occurs with other clitics such as the question clitic \textit{-mI}. Consider \autoref{ex:questionclitic}. When attached, the question clitic comes between the TAM marker and the copula.

\begin{exe}
    \ex \label{ex:questionclitic}
    \gll Bu kitab-{\i} oku-yacak m{\i}=y-d{\i}-n?\\
    this book-\textsc{acc} read-\textsc{fut} \textsc{q}=\textsc{cop}-\textsc{pst}-\textsc{2sg}\\
    \glt `Were you going to read this book?'
\end{exe}

Another clue for the clitic status of the copula is its interaction with vowel harmony. When detached, it has its own phonological domain; thus vowel harmony processes do not percolate from the main verb to the copula as seen in \autoref{ex:VerbalAnalyBe}.

However, semantic contributions of TAM markers and their interaction with each other provides a counterpoint for segmenting the copular clitic.\footnote{We thank the anonymous reviewer for pointing out this issue and initiate this discussion.} On a first look, verbs with a copular clitic seem to carry two different tense information. However, two consecutive TAM markers in Turkish do not imply two tenses. While one of them still provides tense information, the other one implies additional aspect. Consider the verb \emph{gelecektim} in \autoref{ex:acakti}. When either suffix (\textit{-ecek} or \textit{-ti}) is attached to a verb without any additional TAM marker, they mainly provide the tense information. When they are used together as in \autoref{ex:acakti}, the suffix \textit{-ti} implies the tense information, and the suffix \textit{-ecek} provides the prospective aspect information. This aspect of the copular clitic points towards a solution in which verbs with a copular clitic should be analyzed as a single unit.

\begin{exe}
\ex\label{ex:acakti} 
\gll Okul-a gel-ecek=ti-m ama fikri-m-i de\u{g}i\c{s}tir-di-m.\\
school-\textsc{dat} come-\textsc{fut}=\textsc{pst}-\textsc{1sg} but mind-\textsc{poss.1sg}-\textsc{acc} change-\textsc{pst}-\textsc{1sg}\\
\glt `I was going to come to the school, but I changed my mind.'
\end{exe}

After exchanging ideas on this issue within the UD community\footnote{For the whole discussion, see \url{https://github.com/UniversalDependencies/docs/issues/639}} and considering points mentioned in this section, we decided to segment all instances of the copular verb \textit{i-} as a copula (\texttt{cop}). With this change, we unified the treatment of all clitics that may attach to a root which include the question particle \textit{=m{\i}}, focus particles like \textit{=da}, and copular verb particles; thus, followed the UD dependency relations more faithfully.

\subsubsection{Compound} 

Another inconsistent annotation in the previous Turkish treebanks was compounds and their classification. The UD framework suggests that \texttt{compound} should be tailored to each language with its particular morphosyntax. Mostly in Turkish PUD, also in other Turkish UD-treebanks, constituents that carry a morphological marker for possessive-compounds are annotated as \texttt{compound} like in \autoref{ex:compound}. The name `possessive-compounds' is how the linguistic literature refers to it, but for our purposes we take it as a compositional structure and separate it from the UD dependency `compound'. This means that our criteria for compound-hood are syntactic composition properties. We have modified cases with the morphological marker \textit{-(s)I(n)} as \texttt{nmod:poss}, which is already a convention in use in the UD framework.\\

\begin{exe}
    \ex
        \scalebox{1}{
        \begin{dependency}
            \begin{deptext}[column sep=0.3cm]
            Bun-lar-{\i}n \& elli-si \& pazar \& alan-{\i} \& =y-d{\i} \& . \\
            this-\textsc{pl}-\textsc{gen} \&  fifty-\textsc{poss} \& market \& place-\textsc{poss} \& =\textsc{cop}-\textsc{pst} \& . \\
        \end{deptext}
            \deproot[edge unit distance=2ex]{4}{\textsc{root}}
            \depedge[edge unit distance=2.5ex]{4}{2}{\textsc{nsubj}}
            \depedge{2}{1}{\textsc{nmod:poss}}
            \depedge{4}{5}{\textsc{cop}}
            \depedge{4}{6}{\textsc{punct}}
            \depedge[edge unit distance=3ex,edge style={ultra thick}]{4}{3}{\textsc{\textbf{nmod:poss}}}
            \depedge[edge unit distance=2ex,edge below, label style={below},edge style={dotted}]{4}{3}{\textsc{compound}}
        \end{dependency}
        }
    \glt `50 of these were marketplaces.'
    \label{ex:compound}
\end{exe}

Turkish employs different strategies for compounding. These strategies can display differences in their morphological and phonological forms. For our purposes, we divide them into two: (i) compounds with the compound marker \textit{-(s)I(n)} and (ii) compounds without the compound marker \textit{-(s)I(n)}. Some compound types without the compound marker are given in \autoref{ex:compounds}. These compounds are formed with different types of lexical inputs and can have varying degrees of morpho-phonological properties, none of which employs a compound marker. We annotated the compounds that do not employ a marker as \texttt{compound}.
\begin{exe}
    \ex \label{ex:compounds}
\begin{multicols}{3}    
    \begin{xlist}
    \ex Noun + Noun
    \gll \c{s}i\c{s} kebap \\ skewer kebab \\ \glt `shish kebab' 
     
    \ex Non-word + Non-word
    \gll abur cubur \\ {\ldots} {\ldots} \\ \glt `junk food'
     
    \ex Noun + Non-Word
    \gll kitap mitap \\ book \textsc{emph}\textasciitilde book \\ \glt `book and whatnot'
   
    \ex Adverb + Adverb
    \gll bug{\"{u}}n yar{\i}n \\ today tomorrow \\ \glt `soon'

    \ex Verb + Verb 
    \gll in-di bin-di \\ on-\textsc{pst} off-\textsc{pst} \\ \glt `stopover'
     
    \ex Adjective + Adjective
    \gll k{\i}r{\i}k d{\"{o}}k{\"{u}}k \\ broken dowdy \\ \glt `scrap'
    \end{xlist}
\end{multicols}    
\end{exe}

The important distinction for our purposes is the existence of the compound marker \textit{-(s)I(n)}. This marker is only observed in Noun+Noun compounds and most of these compounds can be turned into Genitive-Possessive constructions as in \autoref{ex:compounds2}.

\begin{exe}
    \ex \begin{multicols}{2}  \label{ex:compounds2} 
    \begin{xlist}
     \ex  Noun + Noun 
     \gll okul bina-s{\i} \\ school building-\textsc{3sg} \\ \glt `school building'
\columnbreak     
     \ex Possessive construction
     \gll okul-un bina-s{\i} \\ school-\textsc{gen} building-\textsc{3sg} \\ \glt `the school's building'
     \end{xlist}
    \end{multicols} 
\end{exe}

We annotated Noun+Noun compounds that employ the compound marker \textit{-(s)I(n)} as \texttt{nmod:poss}. There are three reasons behind this decision. The first one is that the marker does not survive in possessive constructions, it is replaced by the possessive markers. If the possessor is \textsc{1sg} or \textsc{2sg}, the marker is replaced with first person singular possessive \textit{-(I)m} or the second person singular possessive \textit{-(I)n}, respectively. If the possessor is \textsc{3sg} the marker stays the same. The second reason is plural marking of the compounds. Any plural marking precedes the marker \textit{-(s)I(n)} as opposed to following it, just like in possessive constructions (\autoref{compoundposs}). The third reason is that compounds formed with the marker \textit{-(s)I(n)} can have their modifier (non-head) be subject to questions, whereas compounds without it cannot (\autoref{compextraction}). Questions are considered to be extractions out of syntactic structures which can not target parts of a word form.
\begin{exe}
\ex \begin{multicols}{2} \label{compoundposs}
\begin{xlist}
    \ex \gll ders kitap-(lar)-{\i} \\ course book-\textsc{pl}-\textsc{3sg} \\ \glt `coursebook(s)'
    \ex \gll ders kitap-(lar)-{\i}m \\ course book-\textsc{pl}-\textsc{1sg} \\ \glt `my coursebook(s)'
\columnbreak    
    \ex \gll ders kitap-(lar)-{\i}n \\ course book-\textsc{pl}-\textsc{2sg} \\ \glt `your coursebook(s)'
    \ex \gll (o-nun) ders kitap-(lar)-{\i} \\ (s/he-\textsc{gen}) course book-\textsc{pl}-\textsc{3sg} \\ \glt `his/her coursebook(s)'
\end{xlist}
\end{multicols}

\ex \label{compextraction}
\begin{xlist}
\ex \begin{xlisti}
\ex \gll adana kebap \\ A kebab \\ \glt `Adana kebab'
\ex \gll *Ne kebap ye-di? \\ what kebab eat-\textsc{pst}[\textsc{3sg}] \\ \glt Intended `What type of kebab did (s/he) eat?'
\end{xlisti}

\ex \begin{xlisti}
\ex \gll adana kebab-{\i} \\ A kebab-\textsc{3sg} \\ \glt `Adana kebab'
\ex \gll Ne kebab-{\i} ye-di? \\ what kebab-\textsc{3sg} eat-\textsc{pst}[\textsc{3sg}] \\ \glt `What type of kebab did (s/he) eat?'
\end{xlisti}
\end{xlist}

\end{exe}

As a result, (i) the marker \textit{-(s)I(n)} not surviving possessive constructions and the ability to transition from a compound to genitive-possessive construction shows that the marker \emph{-(s)I(n)} and possessive markers are in a disjunctive blocking relation. This suggests that they are competing for similar grammatical functions. (ii) The plural marker linearizes before the marker \textit{-(s)I(n)}. If \textit{-(s)I(n)} was part of the word form, the plural marking should have linearized to the right of it. This shows that the marker \textit{-(s)I(n)} is not part of the word form. (iii) Parts of the construction formed by \textit{-(s)I(n)} can be targeted by questions. Question formations only target syntactic constituents and not part of word forms. This indicates that structures with \textit{-(s)I(n)} do not constitute an indivisible word form. All these three reasons make constructions involving \textit{-(s)I(n)} more syntactic (compositional) than morphological. This does not unilaterally rule out the constructions with \textit{-(s)I(n)} as compounds, but within the framework of UD they are more suited to be classified as \texttt{nmod:poss} than \texttt{compound}.

There is a robust linguistics discussion about the status of the marker \textit{-(s)I(n)} as being classified either as a compound or as an agreement marker. The word forms produced by it are actually referred to as `possessive compounds' \citep{hayasi1996dual,kunduraci2013turkish,taylanozturk2014,ozturk2016possessive}, introducing a dilemma even in its own name.

\subsubsection{Classifier}


The use of the classifier syntactic dependency (\texttt{clf}) was also inconsistent within the already existing Turkish UD treebanks. In the UD guidelines, the use of \texttt{clf} is limited to languages with highly grammaticized classifier systems. The difference between classifier languages and non-classifier languages is depicted with Chinese (classifier) and English (non-classifier). However, this distinction is not always clear-cut in other languages like Turkish \citep{Sag2019}. According to \citet{goksel2005}, numerals can be followed by certain elements such as the enumerator \textit{tane} (piece), measurement denoting words such as \emph{dilim} (slice) and \emph{{\c{s}}i{\c{s}}e} (bottle), and membership/identity denoting words like \emph{{\"{o}}rnek} (example) and \emph{kopya} (copy). They show that even though these elements are optional between a numeral and a noun, in partitive constructions with ablative cases, they are obligatorily used. The examples below show that the classifier \textit{tane} (piece) is optional in sentences like \autoref{ex:usualClassifier}. However, when the classifier is in inflected form, deleting it makes the sentence ungrammatical as in \autoref{ex:partitive}. The sentence becomes marginally acceptable when the inflection is concatenated to the numeral as in \autoref{ex:partitive2}.

\begin{exe}
\ex \label{ex:classifiers}
\begin{xlist}
\ex[\phantom{\%}]{  \label{ex:usualClassifier}
\gll D\"{o}rt (tane) elma al-d{\i}-m. \\ four piece apple buy-\textsc{pst}-\textsc{1sg} \\ \glt `I bought four apples.'}
\ex[\phantom{\%}]{ \label{ex:partitive}
\gll K\"{u}\c{c}\"{u}k-ler-den on *(tane-si) yeter mi? \\
small-\textsc{pl}-\textsc{abl} ten piece-\textsc{poss} enough \textsc{q} \\
\glt `Will ten of the little ones be enough?' }
\ex[\%]{ {\label{ex:partitive2}
\gll K\"{u}\c{c}\"{u}k-ler-den on-u yeter mi? \\
small-\textsc{pl}-\textsc{abl} ten-\textsc{poss} enough \textsc{q} \\
\glt `Will ten of the little ones be enough?'} }
\end{xlist}
\end{exe}

Apart from the Turkish PUD Treebank, no previous Turkish treebank has used the \texttt{clf} syntactic dependency. In the Turkish PUD Treebank, both measure words and enumerators are annotated using \texttt{clf} dependency. As for the other Turkic treebanks, a measure word \textit{b{\"{o}}telke} (bottle)  in the Kazakh UD Treebank is annotated using \texttt{clf}. On the other hand, in the Uyghur UD Treebank, no \texttt{clf} is used. In addition to the UD Treebanks, other recent treebanks such as \citet{Kayadelen2020} that use dependency grammar framework in their annotation, make use of the classifier dependency relation for both enumerators and measurement denoting words. 

In the BOUN Treebank and our re-annotated versions of PUD and IMST-UD, we annotated enumerators like \textit{tane} (piece) and \textit{adet} (piece) as classifiers and used the \texttt{clf} dependency relation.  A slightly modified example sentence from our treebank can be seen in \autoref{ex:classifiers3}. One of the UD framework's core ideas is to create a typologically comparable set of treebanks. In this direction, it is important to reflect the use of classifier words in Turkish, even if they are optional.

\begin{exe}
    \ex
        \scalebox{1}{
        \begin{dependency}
            \begin{deptext}[column sep=0.3cm]
            \"{U}\c{c} \& adet \& yumurta-y{\i} \& kar{\i}\c{s}-t{\i}r-{\i}n \& . \\
            three \&  \textsc{cl} \& egg-\textsc{acc} \& mix-\textsc{caus}-\textsc{imp.2sg} \& . \\
        \end{deptext}
            \deproot{4}{\textsc{root}}
            \depedge{4}{3}{\textsc{obj}}
            \depedge{3}{1}{\textsc{nummod}}
            \depedge{1}{2}{\textsc{clf}}
            \depedge{4}{5}{\textsc{punct}}
        \end{dependency}
        }
    \glt `Mix three eggs.'
    \label{ex:classifiers3}
\end{exe}

\subsubsection{Core Arguments} 

Turkish also poses a problem with regards to the detection of core arguments. This problem stems from mainly two reasons: core arguments marked with a lexical case and object drop of the core arguments. Like Czech, Turkish allows its direct object to be marked with oblique cases. In addition to the structural accusative case, Turkish also makes use of dative, ablative, comitative and locative on objects, which are the cases that adjuncts can also take. Both the adjunct in \autoref{ex:adjcase} and the core argument in \autoref{ex:lexcase} are marked with the same case: \textsc{com} (comitative). When there is no appropriate context that introduces the object earlier, a \textsc{com}-marked NP becomes obligatory as in \autoref{ex:lexcase}. However, \autoref{ex:adjcase} is completely fine regardless of the context and the existence of the \textsc{com}-marked NP. This is because the \textsc{com}-marked NP is a core argument in \autoref{ex:lexcase}, whereas it is an adjunct in \autoref{ex:adjcase}.\footnote{Note that in certain environments where there is an immediate follow-up sentence to \autoref{ex:lexcase}, \textsc{com}-marked argument can still be omitted as in (\ref{ex:lexcase2}). We thank the anonymous reviewer for pointing this out.
\begin{exe}
\ex \label{ex:lexcase2}
\gll Serap hep {dalga ge\c{c}}-er, ama karde\c{s}-i hi\c{c} k{\i}z-ma-z-d{\i}.\\
Serap always make.fun-\textsc{aor} but sibling-\textsc{poss} never get.angry-\textsc{neg}-\textsc{aor.neg}-\textsc{pst}\\
\glt `Serap would always make fun of her sister and she would never get angry.'
\end{exe}

} As it can be seen from the examples, Turkish can drop its object without any marking on the verb when it is available in the discourse or it is not contradictory within a given context. Since it is impossible to drop the new information or correction in the case of \autoref{ex:lexcase} without a context that introduces the direct object earlier, we conclude that the NP \textit{k{\i}z karde\c{s}iyle} (with her sister) is a core argument. If it were just an adjunct, the phrase can be omittable. 

\begin{exe}
    \ex \label{ex:lexcase}
    \gll Serap *(k{\i}z karde{\c{s}}-i-yle) hep {dalga ge\c{c}}-er.\\
    Serap {\hspace{.7em}girl} sibling-\textsc{poss}-\textsc{com} always make.fun-\textsc{aor}. \\
    \glt `Serap always makes fun of her sister.'
    \ex \label{ex:adjcase}
    \gll Serap okul-a (abla-s{\i}-yla) gid-er.\\
    Serap school-\textsc{dat} big.sister-\textsc{poss}-\textsc{com} go-\textsc{aor}\\
    \glt `Serap goes to school (with her elder sister).'
\end{exe}

Oblique case marking of the core arguments together with the optionality of the contextually available core arguments yields a problem for 
the annotation process within a framework where the difference between core arguments and non-core arguments is a morphologi-cally-apparent case marking as in the UD framework. Recent discussions in the UD framework also acknowledge this problem \citep{zeman2017core,przepiorkowski2018arguments}. They propose a new dependency relation: \texttt{obl:arg}. In our annotations, we used the \texttt{obj} dependency relation as in \autoref{ex:corearg}. The UD guidelines state that even though \texttt{obj} often carries an accusative case, it may surface with different case markers when the verb dictates a different form, in our case \textit{lexical} cases like \textsc{com} (\autoref{ex:lexcase}) and \textsc{abl} (\autoref{ex:corearg}). This approach is also utilized within the most recent Turkish treebank in which they did not distinguish between the objects with accusative case and the objects with non-accusative cases \citep{Kayadelen2020}.

\begin{exe}
\ex 
        \scalebox{1}{
        \begin{dependency}
            \begin{deptext}
            {\"U}t{\"u}-den \& anla-ma-m \&. \\
            ironing-\textsc{abl} \& understand-\textsc{neg}-\textsc{1sg} \& . \\
        \end{deptext}
            \deproot{2}{\textsc{root}}
            \depedge[edge style={ultra thick}]{2}{1}{\textsc{\textbf{obj}}}
            \depedge[edge unit distance=4ex,edge style={densely dotted}]{2}{1}{\textsc{obl}}
            \depedge{2}{3}{\textsc{punct}}
        \end{dependency}
        }
    \glt `I do not know anything about ironing.'
    \label{ex:corearg}
\end{exe}

Another core argument specified in the UD guidelines is the \texttt{iobj} argument. In their assessment of Turkic treebanks, \citet{Tyers2017} suggest using case promotion or demotion in passivization or causativization as a clue for determining argumenthood. When sentences are passivized in Turkish, the structural case accusative on the object is deleted in the transformation whereas oblique cases such as the ablative case is not deleted. They use this asymmetry to argue for a non-core analysis of oblique case marked objects. In their proposed annotation scheme, only tokens with non-oblique cases should be annotated as a core argument since only non-oblique cases go through case promotion or demotion. However, as we have previously shown in this section, objects marked with oblique cases behave the same as the objects marked with the accusative cases. Turkish can have oblique cases as a marker of objects even though they do not go through case demotion in passive sentences as in \autoref{ex:imppasive}.

\begin{exe}
\ex \label{ex:imppasive}
\gll \"{U}t\"{u}-den de anla-n-ma-z m{\i}?\\
ironing-\textsc{abl} \textsc{emp} understand-\textsc{pass}-\textsc{neg}-\textsc{3sg} \textsc{q}?\\
\glt `How can one not know anything about ironing?'
\end{exe}

Following the reasons specified in this section, we did not make use of case clues in the annotation of \texttt{iobj}, instead we utilized the effects born out of context. Following our annotation process, we should annotate the dative marked noun \textit{bana} (to me) using the \texttt{iobj} dependency relation if we cannot omit it when the information is already available in the discourse. Without any existing prior context, one cannot omit the dative marked noun in sentences like \autoref{ex:iobj} where the main predicate is ditransitive. 

\begin{exe}
\ex 
        \scalebox{1}{
        \begin{dependency}
            \begin{deptext}
            Deniz \& kitab-{\i} \& ban-a \& ver-di \&. \\
            Deniz \& book-\textsc{acc} \& \textsc{1sg}-\textsc{dat} \& give-\textsc{pst}\\
        \end{deptext}
            \deproot{4}{\textsc{root}}
            \depedge{4}{3}{\textsc{iobj}}
            \depedge{4}{2}{\textsc{obj}}
            \depedge{4}{1}{\textsc{nsubj}}
        \end{dependency}
        }
    \glt `Deniz gave me the book.'
    \label{ex:iobj}
\end{exe}

In addition to our treebank, the \texttt{iobj} dependency relation is also used in other Turkish and Turkic treebanks. Prior to our re-annotation, the Turkish PUD Treebank already made use of this dependency relation. With our re-annotation, the IMST-UD Treebank also utilizes the \texttt{iobj} dependency. 
The \texttt{iobj} relation is also used in a Turkic treebank: the UD Kazakh Treebank \citep{Tyers2017,makazhan_tl2015}. We believe that the non-optionality of cases like \emph{bana} (to me) in \autoref{ex:iobj} and its already existing use in other Turkish and Turkic treebanks justify our usage as well.

\subsubsection{Summary of the linguistic considerations}

The points made through the linguistic considerations are based on the idea that a language phenomenon needs to be evaluated with regards to its interactions with other phenomena in the same language. There could be opaque processes which require referring to the derivational history of a construction such as nominalization in embeddings, argument dropping (subject, object, indirect object), compound making strategies, or grammatical functions of a clitic. Additionally, a language does not need to employ a structural property uniformly in its grammatical system. Classifiers in Turkish could be an example for this. Example sentences for the UD tagset could already exist in the provided guidelines, but they lack linguistic diagnostics which are crucial to differentiate between the closely related constructions and the mostly opaque processes in a given language. We hope explicitly stating the diagnostics used for an annotation scheme becomes a practice so that the unification process of the treebanks does not follow from standalone examples but rather from testable predictions.

\section{Annotation Tool} \label{sec:tool}

Annotation tools are fundamental to the facilitation of the annotation process of many NLP tasks including dependency parsing. UD treebanks are re-annotated or annotated from scratch in line with the annotation guidelines of the UD framework \citep{nivre2016universal}. There are several annotation tools that are showcased within the UD framework such as UD Annotatrix \citep{tyers-etal-2017-ud} and ConlluEditor \citep{heinecke2019conllueditor}. These tools are mostly based on mouse-clicks, and provide graph view and/or text view. Morphological features are, in general, not easy to annotate/edit with the available tools. There are also annotation tools that have been developed for annotating Turkish treebanks \citep{Atalay2003, yildiz2016constructing, eryiugit2007itu, Pamay2015}. However, they are not specific to the UD framework. Apart from that, they do not have practical user interfaces regarding dependency parsing.

We present BoAT, a new annotation tool specifically designed for dependency parsing. To the best of our knowledge, it is the first tool that provides tree view and table view simultaneously. BoAT enables annotators to use both mouse clicks and keyboard shortcuts. In addition, unlike previous dependency parsing annotation tools which show morphological features as a whole, in BoAT, morphological features are parsed and expanded into multiple columns, as they are one of the most re-annotated fields according to the observations of our annotators. The enhanced presentation of morphological features is beneficial for annotators. Using BoAT, tokenization can be easily changed by splitting or joining tokens. This is a useful property, especially for agglutinative languages since they have more suffixes, and tokenization may differ according to the used methods. The tool itself, however, is not specific to agglutinative languages and can be used for other languages as well.

BoAT is designed with the aim of presenting a user-friendly, compact, and practical manual annotation tool that is built upon the preferences of the annotators. It combines useful features from other tools such as changing the tokenization, using a validation mechanism, and taking notes with novel features such as combining tree and table views, parsing morphological features, and adding keyboard shortcuts to match the needs of the annotators for the dependency parsing task. 

While developing BoAT, we received feedback from our annotators in every step of the process. One crucial aspect of annotation is speed. Annotation tools are helpful in this regard but they are still open to advancement in terms of speed. The existing tools  within the UD framework mostly rely on mouse clicks and dragging, and the usage of keyboard shortcuts is very limited. Unlike them, almost every possible action within BoAT can be carried out via both mouse clicks and keyboard shortcuts. We aim to decrease the time-wise and ergonomic load introduced by the use of a mouse and to increase speed accordingly.

We also added the note taking option being inspired by BRAT \citep{stenetorp-etal-2012-brat}. While notes are specific to annotations in BRAT, they are specific to each sentence in our tool. This feature enabled our annotators to have better communication and have better reporting power.

\subsection{Features}

\begin{figure}[hbt!]
    \centering
    \includegraphics[width=0.9\textwidth]{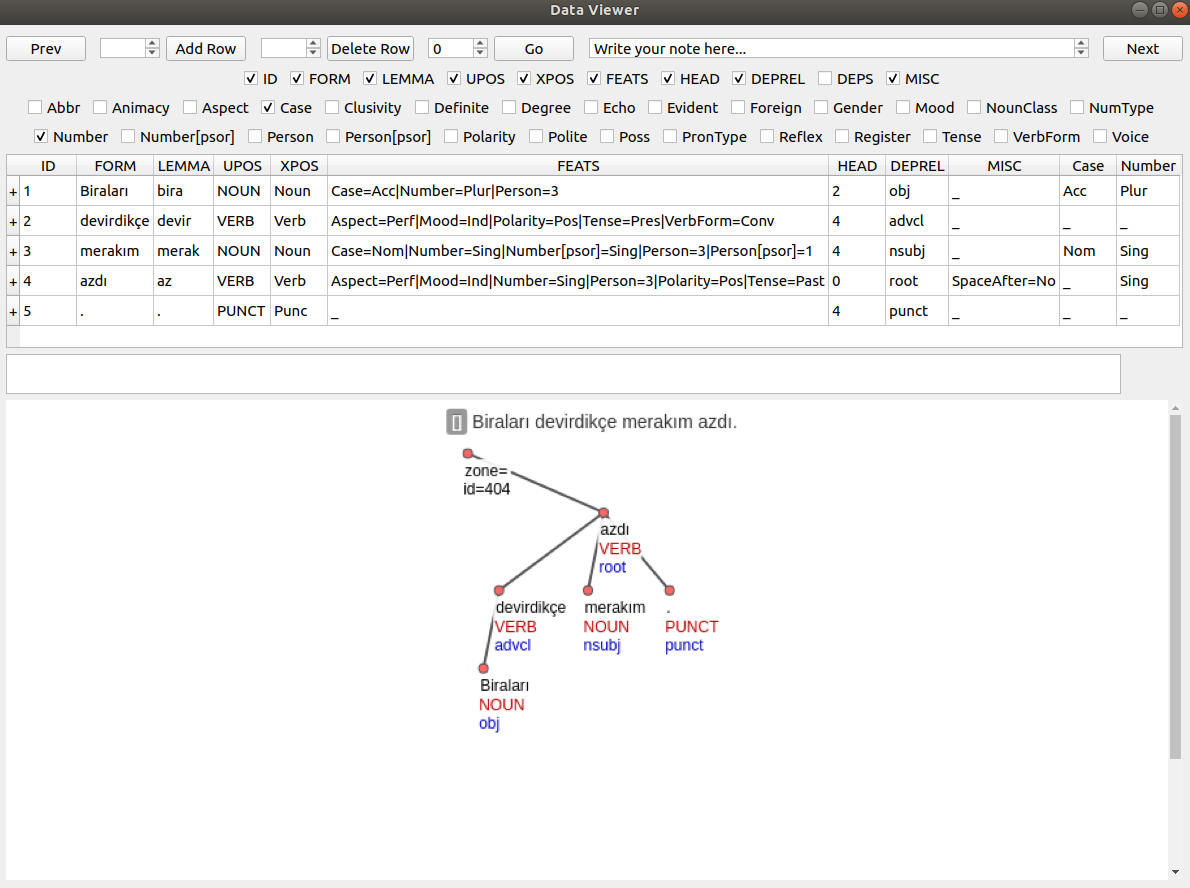}
    \caption{A screenshot from the tool. The sentence is taken from \autoref{ex:nmodtoadvcl}.}
    \label{fig:tool}
\end{figure}

BoAT is a desktop annotation tool which is specifically designed for CoNLL-U files. It offers both tree view and table view as shown in \autoref{fig:tool} for an example sentence. The upper part of the screen shows the default table view while the lower part shows the tree view. Below we explain briefly the components and some of the properties of the tool.

\textbf{Tree view:} The dependency tree of each sentence is visualized in the form of a graph. Instead of using flat view, hierarchical tree view is used. If the user hovers the mouse pointer over a token in the tree, the corresponding token in the sentence above the tree is highlighted which gives the user a linearly readable tree in order to increase readability and clarity. The tree view is based on the hierarchical view feature in the CoNLL-U Viewer offered by the UD framework. 

\textbf{Table view:} Each sentence is shown along with its default fields which are ID, FORM, LEMMA, UPOS, XPOS, FEATS, HEAD, DEPREL, DEPS, and MISC. The morphological features denoted by the FEATS field are parsed into specific subfields. These subfields are a subset of universal and language-specific features in the UD framework. These subfields are optional in the table view; annotators can choose which subfields they want to see. They are stored in the CoNLL-U file in a concatenated manner.

\textbf{Customizing table view:} Annotators can customize the table view according to their needs by using the checkboxes assigned to the fields and the subfields of the FEATS field shown above the parsed sentence. In this way, the user can organize the table view easily and obtain a clean view by removing the unnecessary fields when annotating. This customization ameliorates readability, and consequently the speed of the annotation. The example in \autoref{fig:tool} shows a customized table view.

\textbf{Actions in table view:} To ease the annotation process, the most frequently used functions are assigned to keyboard shortcuts. Moreover, annotators can jump to any sentence by simply typing the ID of the sentence. The value in a cell is edited by directly typing when the focus is on that cell. If one of the features is edited, the FEATS cell is updated accordingly. 

\textbf{Changing tokenization:} One of the biggest challenges in the annotation process is keeping track of the changes in the segment IDs when new segmentations are introduced. In BoAT, new tokens can be added or existing ones can be deleted to overcome tokenization problems generated during the pre-processing of the text. Moreover, annotating multiword expressions often comes at the cost of updating the segment IDs within a sentence in the case of misdetected multiword expressions due to faulty automatic tokenization. Annotators may need an easy way to split a word into two different units. We enabled our annotators to split or join words within our tool by clicking the cells in the first column of the table (written ``+” or ``-”) or using keyboard shortcuts, which permits a more accurate analysis of multiword expressions.

\textbf{Validation:} Each tree is validated with respect to the field values before saving the sentence. If an error is detected in the annotated sentence, an error message is issued such as ``unknown UPOS value". The error is shown between the table view and the tree view.  

\textbf{Taking notes:} With the note feature, the annotator is able to take notes for each sentence as exemplified on the topmost line in \autoref{fig:tool}. Each note is attached to the corresponding sentence and stored in a different file with the ID of the sentence.

\subsection{Implementation}

BoAT\footnote{BoAT is available at \url{https://github.com/boun-tabi/BoAT}} is an open-source desktop application. The software is implemented in Python 3 along with PySide2 and regex modules. In addition, CoNLL-U viewer is utilized by adapting some part of the UDAPI library \citep{popel-etal-2017-udapi}. Resources consisting of a data folder, the tree view, and validate.py are adapted from the UD-maintained tools\footnote{\url{https://github.com/universaldependencies/tools}} for validation check. The data folder is used without any changes while some modifications have been made to validate.py. BoAT is a cross-platform application since it runs on Linux, OS X, and Windows. 

The BoAT tool was designed in accordance with the needs of the annotators, and it increases the speed and the consistency of the annotation process on the basis of our annotators' feedbacks. Currently, BoAT only supports the ConLL-U format of UD since it was designed specifically for dependency parsing. In the future, it may be extended to support other formats such as the ConLL-U Plus format.\footnote{\url{https://universaldependencies.org/ext-format.html}}

\section{Experiments} \label{sec:exp}

We report the results of our parsing experiments on the BOUN Treebank as well as on its different text types, which will serve as a baseline for future studies. In addition to the brand-new BOUN Treebank, we performed parsing experiments on our re-annotated versions of the IMST-UD \citep{turk-etal-2019-improving} and PUD \citep{turk-etal-2019-turkish} treebanks,\footnote{These treebanks are available at \url{https://github.com/boun-tabi/UD_Turkish-BIMST} and \url{https://github.com/UniversalDependencies/UD_Turkish-PUD}} in order to observe the effect of using additional training and test data.

Most prior studies \citep{eryiugit2008dependency, hall2010single, el2014initial, sulubacak2016imst, sulubacak-etal-2016-universal, sulubacak2018implementing} on Turkish dependency parsing evaluate the treebanks they use (mostly versions of the IMST-UD Treebank) using MaltParser \citep{nivre2007maltparser}. However, the definition of a well-formed dependency tree for MaltParser is different than the conventions of UD such that the root node may have more than one child in the output of the MaltParser. UD defines a dependency tree with exactly one root node, and it is not possible to have MaltParser produce dependency trees that follow the UD convention. For this reason, we use the Stanford's neural parser whose original version \citep{dozat2017stanford} achieved the best parsing scores on the IMST-UD Treebank with 69.62 UAS and 62.79 LAS in the CoNLL 2017 Shared Task on Multilingual Dependency Parsing from Raw Text to Universal Dependencies \citep{zeman-EtAl:2017:K17-3}, and its modified version \citep{kanerva-EtAl:2018:K18-2} achieved one of the best performances on the same treebank with 70.61 UAS and 64.79 LAS in the follow-up task in 2018 \citep{zeman-EtAl:2018:K18-2}. It is currently one of the state-of-the-art dependency parsers. This parser uses unidirectional LSTM modules to generate word embeddings and bidirectional LSTM modules to create possible head-dependency relations. It uses ReLu layers and biaffine classifiers to score these relations. For more information, see \citet{dozat2017stanford}.

As stated in Section \ref{sec:treebank}, the BOUN Treebank consists of 9,761 sentences from five different text types. These text types almost equally contribute to the total number of sentences. For the parsing experiments, we randomly assigned each section to the training, development, and test sets with the 80\%, 10\%, and 10\% percentages, respectively. \autoref{tnc-division} shows the number of sentences in each set of the BOUN Treebank.

\begin{table}[hbt!]
\centering
\caption{Division of the BOUN Treebank and its different sections 
among training, development, and test sets for the experiments.}
\begin{tabular}{cc}
\begin{tabular}{lrrrr}
\hline\noalign{\smallskip}
\bf Treebank & \bf Training & \bf Dev. & \bf Test  & \bf Total \\
\noalign{\smallskip}\hline\noalign{\smallskip}
Essays & 1,561 & 196 & 196 & \bf 1,953\\
Newspapers & 1,518 & 190 & 190 & \bf 1,898\\
Instructional Texts & 1,580 & 198 & 198 &\bf 1,976\\
Popular Culture Articles & 1,568 & 197 & 197 &\bf 1,962\\
Biographical Texts & 1,576 & 198 & 198 & \bf 1,972\\
\hline
\bf BOUN  & 7,803 & 979 & 979 & \bf 9,761\\
\hline

\end{tabular}
\end{tabular}

\label{tnc-division}
\end{table}

In order to observe the parsing performance for different types of text, we first evaluated the dependency parser for each section separately. Then, we measured the performance of the parser on parsing the entire BOUN Treebank. As a final set of experiments, we trained the parser on the training sets of the BOUN Treebank and the re-annotated version of the IMST-UD Treebank separately and together, and tested them on five different settings. With that set of experiments, we aim to measure the difference in performance between the BOUN Treebank and the IMST-UD Treebank and to observe the effect of increasing the training data size on performance for Turkish dependency parsing.

In our experiments, we did not perform pre-processing actions such as removing the sentences from the training or test sets that include non-projective\footnote{In a non-projective sentence, the dependency edges cannot be drawn in the plane above the sentence without any two edges crossing each other, as in (\ref{nonprojective}). However, in a projective sentence, the dependency edges can be drawn in this manner with no edges crossing, as in (\ref{projective}) \citep{nivre2009non}. 
\\
\begin{multicols}{2}
\begin{exe}
\ex  
        \begin{dependency}
            \begin{deptext}
            Siyah \& kedi \& nihayet \& s\"{u}t-\"{u} \& i\c{c}-ti \\
            black \& cat \& finally \& milk-\textsc{acc} \& drink-\textsc{pst}\\
        \end{deptext}
            \deproot{5}{\textsc{root}}
            \depedge{5}{4}{\textsc{\textbf{obj}}}
            \depedge{5}{3}{\textsc{\textbf{advmod}}}
            \depedge{5}{2}{\textsc{\textbf{nsubj}}}
            \depedge{2}{1}{\textsc{\textbf{amod}}}
        \label{projective}
        \end{dependency}
    \glt `Black cat finally drank the milk'
    \columnbreak
\ex  
        \resizebox{\linewidth}{!}{\begin{dependency}
            \begin{deptext}
            Ses-in-i \& sev-iyor-um \& ya\u{g}mur-un \& ben \\
            sound-\textsc{poss}-\textsc{acc}\& love-\textsc{prog}-\textsc{1sg}\& rain-\textsc{gen} \& I \\
        \end{deptext}
        \deproot[edge unit distance=2.5ex]{2}{\textsc{root}}
        \depedge[edge unit distance=4ex]{2}{1}{\textsc{obj}}
        \depedge[edge unit distance=1ex, color=red]{1}{3}{{\color{red}\textsc{nmod:poss}}}
        \depedge[edge unit distance=2ex]{2}{4}{\textsc{nsubj}}
        \label{nonprojective}
        \end{dependency}}
    \glt `I love the sound of the rain'
\end{exe}
\end{multicols}} 
dependencies. All sentences in the treebanks were included in the experiments.  
As for the pre-trained word vectors used by the dependency parser, we used the Turkish word vectors supplied by the CoNLL-17 organization \citep{ginter2017conll}.

For the evaluation of the dependency parser, we used the unlabeled attachment score (UAS) and labeled attachment score (LAS) metrics. UAS is measured as the percentage of words that are attached to the correct head, and LAS is defined as the percentage of words that are attached to the correct head with the correct dependency type. In the experiments, we used gold POS tags instead of automatic predictions of them.

\subsection{Parsing Results on the BOUN Treebank}

\autoref{tnc-res} shows the  parsing results of the test sets for each section in the BOUN Treebank and the BOUN Treebank as a whole in terms of the labeled and unlabeled attachment scores. In these experiments, the parser has been trained by using the entire training set of the BOUN Treebank.

\begin{table}[hbt!]
\centering
\caption{UAS and LAS F1 scores of the parser on the BOUN Treebank.}
\begin{tabular}{cc}
\begin{tabular}{lrr}
\hline\noalign{\smallskip}
\bf Treebank & \bf UAS F1-score &\bf LAS F1-score\\ 
\noalign{\smallskip}\hline\noalign{\smallskip}
Essays & 68.73 & 59.18   \\
Broadsheet National Newspapers & \bf 81.59 & \bf 76.04  \\
Instructional Texts & 79.22 & 72.65  \\
Popular Culture Articles & 77.69 & 71.13  \\
Biographical Texts & 80.28 & 73.68  \\
\hline
BOUN Treebank & 77.36 & 70.37 \\
\hline

\end{tabular}
\end{tabular}

\label{tnc-res}
\end{table}

We observed that the highest and lowest LAS were obtained on the {\it Broadsheet National Newspapers} section and the {\it Essays} section of the BOUN Treebank, respectively. The parser achieved more or less similar performance on the remaining three sections.

To understand the possible reasons behind the performance differences between the parsing scores of the five sections of the BOUN Treebank, we compared the sections with respect to the average token count and the average dependency arc length in a sentence. \autoref{fig:stats} shows these statistics for the five sections of the BOUN Treebank. We observed that both the average token count and the average dependency arc length metrics are the highest in the {\it Broadsheet National Newspapers} section. The second highest in both metrics is the {\it Essays} section. The averages for the {\it Instructional Texts}, {\it Popular Culture Articles}, and {\it Biographical Texts} sections are close to each other. 

\begin{figure}[hbt!]

    \begin{tikzpicture}
       \pgfplotsset{grid style={dashed,gray}}
        \begin{axis}[
         title=Average Token Count,
    symbolic x coords={Training,Development,Test},
    legend style={font=\small, at={(0.2,0.-0.2)},anchor=north west,legend columns=2},
    xtick=data,
    nodes near coords align={vertical},
	enlargelimits=+0.25,
	ybar, 
	        yticklabel style={anchor=near yticklabel,left=0.2cm},
    bar width=0.2cm,    
    ]      

\addlegendentry{Essays}
\addplot plot coordinates {(Training,13.89) (Development,13.51) (Test,13.67)};

\addlegendentry{B. National Newspapers}
\addplot plot coordinates {(Training,15.51) (Development,14.79) (Test,15.56)}; 

\addlegendentry{Instructional Texts}
\addplot plot coordinates {(Training,10.34) (Development,10.75) (Test,10.00)};

\addlegendentry{Popular Culture Articles}
\addplot plot coordinates {(Training,10.69) (Development,10.75) (Test,11.12)};

\addlegendentry{Biographical Texts}
\addplot plot coordinates {(Training,12.02) (Development,11.45) (Test,10.98)};

\legend{}
\end{axis}
    \begin{axis}[
        title=Average Dependency Arc Length,
        symbolic x coords={Training,Development,Test},
        xtick=data,
        nodes near coords align={vertical},
	    enlargelimits=+0.25,
	    ybar, 
        bar width=0.2cm,    
        xshift=6.2cm,
        ybar,
                yticklabel style={anchor=near yticklabel,left=0.2cm},
        legend style={font=\small, at={(0.6,-0.1)},anchor=north east,legend columns=3},
        legend cell align={left},
    ]

\addlegendentry{Essays}
\addplot coordinates {(Training,3.46) (Development,3.65) (Test,3.40)};

\addlegendentry{B. National Newspapers}
\addplot plot coordinates {(Training,3.68) (Development,3.63) (Test,3.76)}; 

\addlegendentry{Instructional Texts}
\addplot coordinates {(Training,3.13) (Development,3.29) (Test,2.99)};

\addlegendentry{Popular Culture Articles}
\addplot coordinates {(Training,3.15) (Development,3.09) (Test,3.22)};

\addlegendentry{Biographical Texts}
\addplot plot coordinates {(Training,3.32) (Development,3.16) (Test,3.09)};

\end{axis}
\end{tikzpicture}
\caption{ The average token count and the average dependency arc length in a sentence for the five sections of the BOUN Treebank. }
\label{fig:stats}
\end{figure}

Note that, {\it the average token count} metric, which shows the length of a sentence, and {\it the average dependency arch length} metric, which depicts the distance between the nodes of the dependency relations in a sentence, can sometimes correlate, although not all long sentences include long range dependencies. We anticipate that the higher these two metrics are in a sentence, the harder the task of constructing the dependency tree of that sentence will be. In \autoref{fig:stats}, we observe that all of the sections except the {\it Broadsheet National Newspapers} conform with this hypothesis. However, the {\it Broadsheet National Newspapers}, which has the highest numbers of these metrics holds the best parsing performance in terms of the UAS and LAS metrics. We believe that these high scores in this section are due to the lack of interpersonal differences in writing in journalese and the editorial process behind the journals and magazines. 

\subsection{Parsing Results on Combinations of Treebanks}
In \autoref{res-all}, we present the success rates of the parser trained and tested on different combinations of the three Turkish treebanks: the BOUN Treebank and the re-annotated versions of the IMST-UD and Turkish PUD treebanks. We chose to include only these two treebanks that we re-annotated because we wanted to measure the effect of our unification efforts for Turkish treebanking on the parsing accuracy.

The parser is trained separately on the training sets of the IMST-UD and BOUN treebanks, and then, by combining these two training sets (denoted as BOUN+IMST-UD in the first column of \autoref{res-all}). Originally created for evaluation purposes \citep{zeman-EtAl:2017:K17-3}, the PUD Treebank is not used in the training phase of these experiments due to its smaller size compared to the other two treebanks; instead, it is used as an additional test set in the evaluations.

Five different test sets are provided in the third column of \autoref{res-all}: the test set of the BOUN Treebank (BOUN), the test set of the IMST-UD treebank (IMST-UD), the Turkish PUD Treebank (PUD), the combined test sets of the BOUN and IMST-UD treebanks (BOUN+IMST-UD), and the combined test sets of the BOUN and IMST-UD treebanks and the PUD Treebank (BOUN+IMST-UD+PUD).

Each of the trained models is tested on these five test sets. We observe the following:
\begin{itemize}
\item The parser model trained on the BOUN Treebank outperforms the one trained on IMST-UD by at least 10\% in LAS on the first and third test sets (and \textasciitilde5\% on the fourth and fifth sets). Not surprisingly, the parser trained on IMST-UD performs better on its own test set (the second test set) than the parser model trained on the BOUN Treebank. However, the performance difference here is smaller than the one when these two models are tested on the BOUN Treebank's test set. To make a comparison, the parser trained on BOUN outperforms the parser trained on IMST-UD by \textasciitilde8\% in UAS and by more than 10\% in LAS when tested on the BOUN test set. On the other hand, for the case of the IMST-UD test set, the parser trained on IMST-UD outperforms the parser trained on BOUN by only \textasciitilde2\% in UAS and LAS. Having less amount of training data and a more inconsistent annotation history might be the cause of the inferior performance of the IMST-UD Treebank when compared to the BOUN Treebank.

\item Joining the training sets of the BOUN and IMST-UD treebanks improves  parsing performance in terms of the attachment scores. The increase in the training size resulted in better parsing scores, contributing to the discussion on the correlation between the size of the corpus and the success rates in parsing experiments \citep{foth2014size, ballesteros2012existing}. 

\item The worst results by all the models were obtained on the PUD Treebank used as a test set. The different nature of the PUD Treebank compared to the other Turkish treebanks may have an effect on this performance drop. This treebank includes sentences translated from different languages by professional translators and hence, the sentences have different structures than the sentences of the other two treebanks. This difference in structures is a result of the different environments in which these texts were brewed, namely a living corpus (BOUN and IMST-UD) and well-edited translations (PUD). 

\end{itemize}

\begin{sidewaystable}
\centering
\caption{The performance of the parser on five different test sets according to UAS and LAS metrics. On each test set, the performance of the parser in the following settings is measured: when trained using only the IMST-UD Treebank, when trained using only the BOUN Treebank, and when trained using these two treebanks together. }
\begin{tabular}{cc}
\begin{tabular}{lr|lr|rr}
\hline\noalign{\smallskip}
\bf Training &\bf Training  & \bf Test & \bf Test &\bf UAS  &\bf LAS \\ 
 \bf set & \bf size &\bf set & \bf size & \bf F1-score &\bf  F1-score\\ 
\noalign{\smallskip}\hline\noalign{\smallskip}
\bf IMST-UD & 3,685 & BOUN  & 979 & 69.38 & 58.65 \\
\hline
\bf BOUN & 7,803 & BOUN  & 979 & 77.36 & 70.37 \\
\hline
\bf BOUN+IMST-UD & 11,488 & BOUN & 979 & 77.57 & 70.50 \\
\hline
\hline
\bf IMST-UD & 3,685 & IMST-UD  & 975 & 75.49 & 65.53 \\
\hline
\bf BOUN & 7,803 & IMST-UD  & 975 & 73.63 & 62.92 \\
\hline
\bf BOUN+IMST-UD  & 11,488  & IMST-UD & 975 & 76.86 & 66.79 \\
\hline
\hline
\bf IMST-UD & 3,685 & PUD  & 1,000 & 65.28 & 49.50 \\
\hline
\bf BOUN & 7,803 & PUD  & 1,000 & 72.33 & 59.57 \\
\hline
\bf BOUN+IMST-UD  & 11,488 & PUD  & 1,000 & 72.76 & 60.39 \\
\hline
\hline
\bf IMST-UD& 3,685 & BOUN+IMST-UD  & 1,954 & 71.89 & 61.62 \\
\hline
\bf BOUN & 7,803 & BOUN+IMST-UD & 1,954 & 75.67 & 66.99 \\
\hline
\bf BOUN+IMST-UD& 11,488 & BOUN+IMST-UD  & 1,954 & 77.25 & 68.82 \\
\hline
\hline
\bf IMST-UD& 3,685 & BOUN+IMST-UD+PUD  & 2,954 & 69.03 & 56.37 \\
\hline
\bf BOUN & 7,803 & BOUN+IMST-UD+PUD & 2,954 & 74.22 & 63.78 \\
\hline
\bf BOUN+IMST-UD& 11,488 & BOUN+IMST-UD+PUD  & 2,954 & 75.30 & 65.17 \\
\hline
\hline

\end{tabular}
\end{tabular}

\label{res-all}
\end{sidewaystable}

 In order to investigate the differences in the percentages of certain dependency relations between the treebanks used in the experiments, we present the distribution of the dependency relation types across the previous\footnote{The re-annotation process was performed on the UD 2.3 versions of these treebanks.} as well as the re-annotated versions of the IMST-UD and PUD treebanks, and the BOUN Treebank in \autoref{comp}.

When comparing the BOUN Treebank and the re-annotated version of the IMST-UD Treebank, we observed that the percentages of the \texttt{case}, \texttt{compound}, and \texttt{nmod} types were lower by more than 1\% in the BOUN Treebank. The percentage of the \texttt{root} type was also lower in the BOUN Treebank by almost 2\%, which indicates that the average token count in sentences is higher in this treebank with respect to the re-annotated version of the IMST-UD Treebank. However, the percentage of the \texttt{nmod:poss} type was higher by more than 2\% and the \texttt{obl} type was higher by more than 3\% in the BOUN Treebank. We believe that these differences are due to the text types we utilized. 
Unlike IMST-UD, the BOUN Treebank includes essay and autobiography text types. These types make frequent use of postpositional phrases such as \textit{bana g{\"{o}}re (in my opinion)} or \textit{1920'ye kadar (until 1920)}, which are encoded with \texttt{case} dependency relations.
Additionally, the language is less formal compared to the non-fiction and news text types, which are the main registers that the IMST-UD Treebank incorporates as indicated in the UD Project. This formality difference explains the lower usage of the \texttt{compound} relation type.

When comparing the BOUN Treebank with the re-annotated version of the Turkish PUD Treebank, we observed that the highest percentage difference was for the \texttt{obl} type which is higher in the BOUN Treebank by more than 7\%.  This difference is again a result of using different text types. The Turkish PUD Treebank consists of Wikipedia articles in which the adjuncts are expected to be used less than the text types we utilized. The other relation types whose percentages are higher in BOUN by more than 1\% were the \texttt{root} type which indicates that the average token count is lower in the BOUN Treebank, and the \texttt{conj} type indicating that the BOUN Treebank has more conjunct relations which sometimes increased the complexity of a sentence in terms of dependency parsing. 

In the comparison of the previous and re-annotated versions of the IMST-UD Treebank with respect to the distribution of dependency relation types, we see that the percentages of the \texttt{advmod}, \texttt{cc},  \texttt{ccomp}, and \texttt{nsubj} types increased by approximately 1\% in the re-annotated version. In contrast, the percentage of \texttt{nmod} is reduced by more than 3\% in the re-annotated version. The reason behind this decrease lies in the fact that in the previous version of the treebank, nominalized verbs which behave like  converbs \citep{goksel2005} are considered nominal modifiers. However, these nominalized verbs actually construct embedded clauses and therefore are treated as clausal modifiers in the re-annotated treebank. In addition, the \texttt{obl} percentage decreased by more than 1\% in the re-annotated version.  

The \texttt{vocative} type no longer exists in the re-annotated version and the newly introduced types that are absent in the previous version are the \texttt{advcl, advcl:cond, aux, cc:preconj, clf, dislocated, goeswith, iobj, orphan}, and \texttt{xcomp} relation labels.

When we analyze the differences between the previous and re-annotated versions of the PUD Treebank, we observe that the biggest difference is in the \texttt{compound} relation with a 10\% reduction. On the other hand, the biggest increase in the percentage of a relation is in the \texttt{nmod:poss} relation with a more than 6\% increase in the re-annotated version. This is because in the previous annotation of the PUD Treebank, some constructions that involve genitive-possessive suffixes are marked with the \texttt{compound} dependency label. Such relations 
have been corrected as \texttt{nmod:poss}. Other noteworthy differences are in the \texttt{fixed} and \texttt{xcomp} relations with a more than 1\% decrease and in the \texttt{flat}, \texttt{nsubj}, and \texttt{obl} relations with a more than 1\% increase in the re-annotated treebank.

\begin{table}[hbt!]
\setlength{\tabcolsep}{4pt}
\centering
\scriptsize
\caption{Comparison of the previous and re-annotated versions of the IMST-UD and PUD treebanks, and the BOUN Treebank on the distribution of dependency relation labels. The black numbers represent the counts and the gray numbers show their percentages.}

\begin{tabular}{lrrrrr}
\hline\noalign{\smallskip}
\bf Relation type & \bf IMST-UD & \bf IMST-UD & \bf PUD & \bf PUD & \bf BOUN \\
 & (previous) & (re-annotated) & (previous) & (re-annotated)  & \\
\noalign{\smallskip}\hline\noalign{\smallskip}
\texttt{acl} & 1,455  {\color{gray}(2.5\%)} & 1,538 {\color{gray}(2.65\%)}& - & 515 {\color{gray}(3\%)} & 3,494  {\color{gray}(2.85\%)} \\
\hline\texttt{acl:relcl} & -& -& 514 {\color{gray}(3.04\%)}& - & -\\
\hline\texttt{advcl} & - & 926 {\color{gray}(1.59\%)} & 405 {\color{gray}(2.4\%)} & 435 {\color{gray}(2.6\%)}& 2,595 {\color{gray}(2.12\%)} \\
\hline\texttt{advcl:cond} & - & 110 {\color{gray}(0.19\%)} & - & 13 {\color{gray}(0.07\%)}& 269 {\color{gray}(0.22\%)}\\
\hline\texttt{advmod} & 1,872 {\color{gray}(3.2\%)} & 2,422 {\color{gray}(4.17\%)}& 1,716 {\color{gray}(10.16\%)}&1,624 {\color{gray}(9.6\%)} & 5,278 {\color{gray}(4.31\%)}\\
\hline\texttt{advmod:emph} &  973 {\color{gray}(1.67\%)} & 976 {\color{gray}(1.68\%)} &145 {\color{gray}(0.86\%)} & 143 {\color{gray}(0.8\%)} & 1,724 {\color{gray}(1.41\%)} \\
\hline\texttt{amod} & 3,451 {\color{gray}(5.94\%)} & 3,337 {\color{gray}(5.74\%)} & 1,224 {\color{gray}(7.25\%)}& 1,318 {\color{gray}(7.8\%)}& 7,869 {\color{gray}(6.43\%)} \\
\hline\texttt{appos} & 51 {\color{gray}(0.09\%)} & 136 {\color{gray}(0.23\%)} & 36 {\color{gray}(0.21\%)} &166 {\color{gray}(1\%)} & 506 {\color{gray}(0.41\%)} \\
\hline\texttt{aux} & - & 1 {\color{gray}(0.002\%)} & 21 {\color{gray}(0.12\%)}& 4 {\color{gray}(0.02\%)}& 39 {\color{gray}(0.03\%)}\\
\hline\texttt{aux:q} & 209 {\color{gray}(0.36\%)} & 211 {\color{gray}(0.36\%)}& - & 1 {\color{gray}(0.01\%)}& 269 {\color{gray}(0.22\%)}\\
\hline\texttt{case} & 2,183 {\color{gray}(3.76\%)} & 2,242 {\color{gray}(3.86\%)} & 694 {\color{gray}(4.1\%)} & 697 {\color{gray}(4.1\%)} & 3,290 {\color{gray}(2.69\%)} \\
\hline\texttt{cc} &  870 {\color{gray}(1.5\%)} & 879 {\color{gray}(3.1\%)} & 519 {\color{gray}(3.1\%)} & 520 {\color{gray}(3.1\%)} & 2,800 {\color{gray}(2.29\%)} \\
\hline\texttt{cc:preconj} & - & 3  {\color{gray}(0.005\%)} & 8 {\color{gray}(0.05\%)} & 8 {\color{gray}(0.05\%)} & 134 {\color{gray}(0.11\%)} \\
\hline\texttt{ccomp} &  36 {\color{gray}(0.06\%)} & 626 {\color{gray}(1.08\%)}& 30 {\color{gray}(0.18\%)}& 171 {\color{gray}(1\%)}& 1,512 {\color{gray}(1.24\%)}\\
\hline\texttt{clf} & - & 8 {\color{gray}(0.01\%)}  & 10 {\color{gray}(0.06\%)}& 10 {\color{gray}(0.06\%)} & 122 {\color{gray}(0.1\%)}\\
\hline\texttt{compound} &  2219 {\color{gray}(3.82\%)} & 1,977 {\color{gray}(3.40\%)} & 2012 {\color{gray}(11.91\%)} & 314 {\color{gray}(1.9\%)} & 2,381 {\color{gray}(1.95\%)}\\
\hline\texttt{compound:lvc} & 512 {\color{gray}(0.88\%)} & 522 {\color{gray}(0.90\%)}& - & 186 {\color{gray}(1.1\%)}& 1,218 {\color{gray}(1.0\%)}\\
\hline\texttt{compound:redup}& 199 {\color{gray}(0.34\%)} &219 {\color{gray}(0.37\%)}& - & 9 {\color{gray}(0.05\%)}& 457 {\color{gray}(0.37\%)}\\
\hline\texttt{conj} & 3,718 {\color{gray}(6.40\%)} &3,529 {\color{gray}(6.07\%)} & 640 {\color{gray}(3.79\%)} & 696 {\color{gray}(4.1\%)} & 7,250 {\color{gray}(5.92\%)}\\
\hline\texttt{cop} & 813  {\color{gray}(1.40\%)} &851  {\color{gray}(1.46\%)}& 517 {\color{gray}(3.06\%)}& 496 {\color{gray}(2.9\%)}& 1,289 {\color{gray}(1.05\%)}\\
\hline\texttt{csubj} & 7 {\color{gray}(0.01\%)} & 82 {\color{gray}(0.14\%)} &  115 {\color{gray}(0.68\%)}& 93 {\color{gray}(0.5\%)}& 546 {\color{gray}(0.45\%)}\\
\hline\texttt{dep} & 1 {\color{gray}(0.002\%)} & 1 {\color{gray}(0.002\%)}& 3 {\color{gray}(0.02\%)} & 3 {\color{gray}(0.02\%)} & 9 {\color{gray}(0.01\%)} \\
\hline\texttt{det} & 2,040 {\color{gray}(3.51\%)} & 1,975 {\color{gray}(3.39\%)}& 671 {\color{gray}(3.97\%)} & 680 {\color{gray}(4\%)} & 4,938 {\color{gray}(4.03\%)}\\
\hline\texttt{det:predet} & - & - &10 {\color{gray}(0.06\%)} & 8 {\color{gray}(0.05\%)}& - \\
\hline\texttt{discourse} &154 {\color{gray}(0.27\%)} &150 {\color{gray}(0.26\%)}& 5 {\color{gray}(0.03\%)} & 5 {\color{gray}(0.03\%)}& 381 {\color{gray}(0.31\%)} \\
\hline\texttt{dislocated} & - &20  {\color{gray}(0.03\%)} & 2 {\color{gray}(0.01\%)}& 5 {\color{gray}(0.03\%)} & 28 {\color{gray}(0.02\%)} \\
\hline\texttt{fixed}  &40 {\color{gray}(0.07\%)}&25  {\color{gray}(0.04\%)} &204 {\color{gray}(1.21\%)} & 1 {\color{gray}(0.01\%)} & 12 {\color{gray}(0.01\%)}\\
\hline\texttt{flat} & 910  {\color{gray}(1.57\%)} &902  {\color{gray}(1.55\%)} & 4 {\color{gray}(0.02\%)}& 409 {\color{gray}(2.4\%)}& 2,039 {\color{gray}(1.67\%)}\\
\hline\texttt{flat:name} & - & - & 247 {\color{gray}(1.46\%)}& - & -\\
\hline\texttt{goeswith} & - & 3  {\color{gray}(0.005\%)} & 1 {\color{gray}(0.01\%)} & 1 {\color{gray}(0.01\%)}& 4 {\color{gray}(0.002\%)}\\
\hline\texttt{iobj} & - & 354  {\color{gray}(0.61\%)}& 90 {\color{gray}(0.53\%)} & 138 {\color{gray}(0.8\%)}& 164 {\color{gray}(0.13\%)}\\
\hline\texttt{list} & - & - & - & - & 40 {\color{gray}(0.03\%)}\\
\hline\texttt{mark} & 76 {\color{gray}(0.13\%)} & 86 {\color{gray}(0.15\%)}& 6 {\color{gray}(0.03\%)}  & 5 {\color{gray}(0.03\%)} & 117 {\color{gray}(0.10\%)}\\
\hline\texttt{nmod} & 3,780 {\color{gray}(6.51\%)} & 1,870 {\color{gray}(3.22\%)}&161 {\color{gray}(0.95\%)} & 174 {\color{gray}(1\%)}& 1,371 {\color{gray}(1.12\%)}\\
\hline\texttt{nmod:arg} &-&- & 110 {\color{gray}(0.65\%)} & - & - \\
\hline\texttt{nmod:poss} &3,534 {\color{gray}(6.08\%)} &3,598 {\color{gray}(6.19\%)}&  722 {\color{gray}(4.27) }& 1,881 {\color{gray}(11\%)}& 10,393 {\color{gray}(8.49\%)}\\
\hline\texttt{nsubj} &3,747 {\color{gray}(6.45\%)} &4,430 {\color{gray}(7.63\%)}& 1,023 {\color{gray}(6.05\%)} & 1,238 {\color{gray}(7.3\%)}&8,499 {\color{gray}(6.94\%)}\\
\hline\texttt{nummod} &621 {\color{gray}(1.07\%)} &567 {\color{gray}(0.98\%)}&207 {\color{gray}(1.22\%)} & 263 {\color{gray}(1.6\%)}& 1,568  {\color{gray}(1.28\%)}\\
\hline\texttt{obj} &4,307 {\color{gray}(7.41\%)} &3,743 {\color{gray}(6.44\%)}& 816 {\color{gray}(4.83\%)} & 945 {\color{gray}(5.6\%)}& 7,381 {\color{gray}(6.03\%)} \\
\hline\texttt{obl} &4,444  {\color{gray}(7.65\%)} &3,824  {\color{gray}(6.58\%)}&  148 {\color{gray}(0.88\%)} & 412 {\color{gray}(2.4\%)}& 12,015 {\color{gray}(9.82\%)}\\
\hline\texttt{obl:tmod} &- & - &  232 {\color{gray}(1.37\%)} & - & - \\
\hline\texttt{orphan} & - & 12 {\color{gray}(0.02\%)}&  12 {\color{gray}(0.07\%)} & 8 {\color{gray}(0.05\%)}& 84 {\color{gray}(0.07\%)}\\
\hline\texttt{parataxis} & 11 {\color{gray}(0.02\%)} & 11 {\color{gray}(0.02\%)}& 74 {\color{gray}(0.44\%)} & 15 {\color{gray}(0.09\%)}& 209 {\color{gray}(0.17\%)}\\
\hline\texttt{punct} & 10,228  {\color{gray}(17.61\%)} & 10,257  {\color{gray}(17.65\%)} & 2,150 {\color{gray}(12.72\%)} & 2,148 {\color{gray}(12.7\%)}& 20,116 {\color{gray}(16.44\%)}\\
\hline\texttt{root} & 5,635  {\color{gray}(9.69\%)} & 5,635  {\color{gray}(9.69\%)}& 1,000 {\color{gray}(5.91\%)}& 1,000 {\color{gray}(5.91\%)} & 9,761 {\color{gray}(7.97\%)} \\
\hline\texttt{vocative}& 1 {\color{gray}(0.002\%)} & - & 1 {\color{gray}(0.001\%)}& - & 88 {\color{gray}(0.07\%)}\\
\hline\texttt{xcomp} & - & 39  {\color{gray}(0.07\%)} & 381 {\color{gray}(2.26\%)} & 125 {\color{gray}(0.7\%)}& 125 {\color{gray}(0.1\%)} \\
\hline
\hline
\bf Total & 58,097 & 58,098 & 16,886 & 16,886 & 122,384\\
\hline
\end{tabular}

\label{comp}
\end{table}

\section{Conclusion} \label{sec:conclusion}

In this paper, we presented the largest and the most comprehensive Turkish treebank with 9,761 sentences: the BOUN Treebank. In the treebank, we encoded the surface forms of the sentences, the universal part of speech tags, lemmas, and morphological features for each segment, as well as the syntactic relations between these segments. We explained our annotation methodology in detail. We also gave an overview of other Turkish treebanks. Moreover, we explained our linguistic decisions and annotation scheme that are based on the UD framework. We provided examples for the challenging issues that are present in the BOUN Treebank as well as other treebanks that we re-annotated. Our treebank with 
a history of the changes we applied and our annotation guidelines are provided online.

In addition to such contributions, we provided a description of our annotation tool: BoAT. We explained our motivation for such an initiative in detail. We also provide the tool and the documentation online.

Lastly, we evaluated our new treebank on the task of dependency parsing. 
We reported UAS and LAS F1-scores with regards to specific text types and treebanks. We also showcased the results of the experiments where our new treebank was used with the re-annotated versions of the IMST-UD and PUD treebanks.
All the tools and materials that are presented in this paper are 
available on our webpage \url{https://tabilab.cmpe.boun.edu.tr/boun-pars}.

\section{Acknowledgements}

We are immensely grateful to Prof. Ye{\c{s}}im Aksan and the other members of the Turkish National Corpus Team for their tremendous help in providing us with sentences from the Turkish National Corpus. We are also thankful to the anonymous reviewers from SyntaxFest'19 and LAW XIII, as well as to \c{C}a\u{g}r{\i} \c{C}\"{o}ltekin for his constructive comments on the re-annotation process of the IMST and PUD Treebanks. GEBIP Award of the Turkish Academy of Sciences (to A.O.) is gratefully acknowledged.

\bibliographystyle{spbasic}
\bibliography{Manuscript.bib}

\newpage

\section{Morphological Conversion} \label{appendix:conversion}

\begin{table}[hbt!]
\setlength{\tabcolsep}{2pt}
\centering
\caption{Mappings of morphological features from the notation of \citet{sak2011resources} to the features used in the UD framework.}
\label{table:morphConversion}
\begin{tabular}{rlcrl}
\hline\noalign{\smallskip}
\citeauthor{sak2011resources} & UD & & \citeauthor{sak2011resources} & UD \\
\noalign{\smallskip}
\hline\noalign{\smallskip}
A1sg    & Number=Sing| Person=1             &  & ByDoingSo & VerbForm=Conv| Mood=Imp               \\
A2sg    & Number=Sing| Person=2             &  & Pos       & Polarity=Pos                         \\
A3sg    & Number=Sing| Person=3             &  & Neg       & Polarity=Neg                         \\
A1pl    & Number=Plur| Person=1             &  & Past      & Aspect=Perf| Tense=Past| Evident=Fh    \\
A2pl    & Number=Plur| Person=2             &  & Narr      & Tense=Past| Evident=Nfh               \\
A3pl    & Number=Plur| Person=3             &  & Fut       & Tense=Fut| Aspect=Imp                 \\
P1sg    & Number[psor]=Sing| Person[psor]=1 &  & Aor       & Tense=Aor| Aspect=Hab                 \\
P2sg    & Number[psor]=Sing| Person[psor]=2 &  & Pres      & Tense=Pres| Aspect=Imp                \\
P3sg    & Number[psor]=Sing| Person[psor]=3 &  & Desr      & Mood=Des                             \\
P1pl    & Number[psor]=Plur| Person[psor]=1 &  & Cond      & Mood=Cnd                             \\
P2pl    & Number[psor]=Plur| Person[psor]=2 &  & Neces     & Mood=Nec                             \\
P3pl    & Number[psor]=Plur| Person[psor]=3 &  & Opt       & Mood=Opt                             \\
Abl     & Case=Abl                         &  & Imp       & Mood=Imp                             \\
Acc     & Case=Acc                         &  & Prog1     & Aspect=Prog| Tense=Pres                          \\
Dat     & Case=Dat                         &  & Prog2     & Aspect=Prog| Tense=Pres                          \\
Equ     & Case=Equ                         &  & DemonsP   & PronType=Dem                         \\
Gen     & Case=Gen                         &  & QuesP     & PronType=Ind                         \\
Ins     & Case=Ins                         &  & ReflexP   & PronType=Prs| Reflex=Yes              \\
Loc     & Case=Loc                         &  & PersP     & PronType=Prs                         \\
Nom     & Case=Nom                         &  & QuantP    & PronType=Ind                         \\
Pass    & Voice=Pass                       &  & Card      & NumType=Card                         \\
Caus    & Voice=Cau                        &  & Ord       & NumType=Ord                          \\
Reflex  & Voice=Rfl                        &  & Distrib   & NumType=Dist                         \\
Recip   & Voice=Rcp                        &  & Ratio     & NumType=Frac                         \\
Able    & Mood=Abil                        &  & Range     & NumType=Range                        \\
Repeat  & Mood=Iter                        &  & Inf       & VerbForm=Vnoun                       \\
Hastily & Mood=Rapid                       &  & FutPart   & VerbForm=Part| Tense=Future| Aspect=Imp\\
Almost  & Mood=Pro                         &  & PastPart  & VerbForm=Part| Tense=Past| Aspect=Perf \\
Stay    & Mood=Dur                         &  & PresPart  & VerbForm=Part| Tense=Pres             \\
While   & VerbForm=Conv| Mood=Imp           &  & ${}$          & ${}$                                    \\
\noalign{\smallskip}\hline
\end{tabular}
\end{table}

\newpage

\section{Word Order Statistics of the BOUN Treebank} \label{appendix:wordorder}

\begin{table}[hbt!]
\centering
\makebox[0pt][c]{\parbox{0.9\textwidth}{%
    \begin{minipage}[b]{0.48\hsize}\centering
    \caption{Word order counts and relative percentages of main arguments within the BOUN Treebank when there is no null argument.}
\label{table:wordorder}
        \begin{tabular}{lrr}
\hline\noalign{\smallskip}
  Order & Count & Percentage (\%) \\ 
\noalign{\smallskip}
\hline\noalign{\smallskip}
 SOV & 1456 & 59.53 \\ 
 OVS & 549 & 22.44 \\ 
 VSO & 165 & 6.75 \\ 
 SVO & 144 & 5.89 \\ 
 OSV & 109 & 4.46 \\ 
 VOS &  23 & 0.94 \\ 
\noalign{\smallskip}\hline
\end{tabular}
    \end{minipage}
    \hfill
    \begin{minipage}[b]{0.48\hsize}\centering
\caption{Word order counts and percentages of main arguments within the BOUN Treebank.}
\label{table:wordorder2}
\begin{tabular}{lrr}
\hline\noalign{\smallskip}
  Order & Count & Percentage (\%) \\ 
\noalign{\smallskip}
\hline\noalign{\smallskip}
 OV & 5744 & 37.21 \\ 
 SV & 5416 & 35.09 \\ 
 SOV & 1456 & 9.43 \\ 
 VS & 1116 & 7.23 \\ 
 VO & 714 & 4.63 \\ 
 OVS & 549 & 3.56 \\ 
 VSO & 165 & 1.07 \\ 
 SVO & 144 & 0.93 \\ 
 OSV & 109 & 0.71 \\ 
 VOS &  23 & 0.15 \\ 
\noalign{\smallskip}\hline
\end{tabular}
    \end{minipage}%
}}
\end{table}

\newpage 

\section{TNC Registers} \label{appendix:tncregisters}

\begin{table}[hbt!]
\setlength{\tabcolsep}{2pt}
\centering
\label{table:tncdetails2}
\begin{tabular}{llrrrr}
\hline\noalign{\smallskip}
ID & Section Name                                                                          & N\textsubscript{Words}  & \%       & N\textsubscript{Documents}& \%       \\
\noalign{\smallskip}
\hline\noalign{\smallskip}
1  & Academic prose: Medicine                                                              & 714,46          & 1.44\%  & 145              & 2.91\%  \\
2  & Academic prose: Social, behavioral sciences                                           & 2,892,961       & 5.83\%  & 432              & 8.66\%  \\
3  & Academic prose: Humanities/Arts                                                       & 2,604,645       & 5.24\%  & 354              & 7.09\%  \\
4  & Academic prose: Natural sciences                                                      & 1,236,958       & 2.49\%  & 251              & 5.03\%  \\
5  & Academic prose: Politics, law, education                                              & 3,857,971       & 7.77\%  & 587              & 11.76\% \\
6  & Academic prose: Technology, computing, engineering                                    & 1,653,909       & 3.33\%  & 251              & 5.03\%  \\
7  & Administrative and regulatory texts, in house use                                     & 155,054         & 0.31\%  & 11               & 0.22\%  \\
8  & Print Advertisements                                                                  & 22,311          & 0.04\%  & 164              & 3.29\%  \\
9  & Biographies/Autobiographies                                                           & 2,372,093       & 4.78\%  & 158              & 3.17\%  \\
10 & Commerce\&Finance/Economics                                                           & 2,282,709       & 4.6\%   & 120              & 2.4\%   \\
11 & E-mail                                                                                & 31,316          & 0.06\%  & 261              & 5.23\%  \\
12 & School essays                                                                         & 56,545          & 0.11\%  & 10               & 0.2\%   \\
13 & Essay                                                                                 & 494,747         & 1\%      & 99               & 1.98\%  \\
14 & Excerpts from modern drama scripts                                                    & 655,618         & 1.32\%  & 63               & 1.26\%  \\
15 & Single and multiple author collections of poems                                       & 279,984         & 0.56\%  & 35               & 0.7\%   \\
16 & Novels/short stories                                                                  & 8,271,257       & 16.65\% & 566              & 11.34\% \\
17 & Official/govermental documents/leaflets
 & 594,45          & 1.2\%   & 56               & 1.12\% \\
18 & Instructional texts                                                                   & 305,829         & 0.62\%  & 29               & 0.58\%  \\
19 & Personal letters                                                                      & 105,693         & 0.21\%  & 4                & 0.08\%  \\
20 & Professional/business letters                                                         & 20,092          & 0.04\%  & 1                & 0.02\%  \\
21 & Miscellaneous texts                                                                   & 1,932,821       & 3.89\%  & 108              & 2.16\%  \\
22 & BNN: arts/cultural material                                & 573,701         & 1.16\%  & 43               & 0.86\%  \\
23 & BNN: commerce \& finance                                   & 759,85          & 1.53\%  & 67               & 1.34\%  \\
24 &BNN:miscellaneous material                                & 545,078         & 1.1\%   & 56               & 1.12\%  \\
25 & Broadsheet national newspapers: science material                                      & 378,432         & 0.76\%  & 23               & 0.46\%  \\
26 & BNN: leisure, belief \& thought   & 1,600,828       & 3.22\%  & 114              & 2.28\%  \\
27 & BNN: sports material                                       & 662,518         & 1.33\%  & 57               & 1.14\%  \\
28 & BNN: column                                                & 418,734         & 0.84\%  & 109              & 2.18\%  \\
29 & Non-academic: medical/health matters                                                  & 99,878          & 0.2\%   & 5                & 0.1\%   \\
30 & Non-academic: social \& behavioural sciences                                          & 2,411,122       & 4.85\%  & 87               & 1.74\%  \\
31 & Non-academic/non-fiction: humanities\&arts                                            & 2,644,260       & 5.32\%  & 156              & 3.13\%  \\
32 & Non-academic: natural sciences                                                        & 93,182          & 0.19\%  & 7                & 0.14\%  \\
33 & Non-academic: politics law education                                                  & 4,934,042       & 9.93\%  & 247              & 4.95\%  \\
34 & Non-academic: technology, computing, engineering                                      & 235,169         & 0.47\%  & 9                & 0.18\%  \\
35 & Popular magazines                                                                     & 667,094         & 1.34\%  & 48               & 0.96\%  \\
36 & Religious texts                                                                       & 975,833         & 1.96\%  & 46               & 0.92\%  \\
37 & Planned speech, whether dialogue or monologue                                         & 455,194         & 0.92\%  & 24               & 0.48\%  \\
38 & Forum                                                                                 & 468,038         & 0.94\%  & 68               & 1.36\%  \\
39 & Blog                                                                                  & 1,199,927       & 2.42\%  & 119              & 2.38\%  \\
   & Total                                                                                 & 49,664,303      &          & 4990             &         \\
\noalign{\smallskip}\hline
\end{tabular}
\end{table}

\end{document}